\documentclass[journal ]{new-aiaa}
\usepackage[utf8]{inputenc}

\usepackage{amsthm}
\theoremstyle{definition}
\newtheorem{definition}{Definition}

\usepackage{textcomp}
\usepackage{booktabs}
\usepackage{multirow}
\usepackage{graphicx}
\usepackage{amsmath}
\usepackage[version=4]{mhchem}
\usepackage{siunitx}
\usepackage{longtable,tabularx}
\usepackage{color, colortbl}
\usepackage[colorinlistoftodos]{todonotes}
\usepackage{etoolbox}
\usepackage[noabbrev,capitalise]{cleveref}
\usepackage{acronym}
\usepackage{etoolbox}
\usepackage{graphicx}
\usepackage{caption}
\usepackage{subcaption}

\newtoggle{ext}
\toggletrue{ext}

\iftoggle{ext}{}{}
\newacro{ENav}{Enhanced AutoNav}
\newacro{QP}{quadratic program}
\newacro{SOCP}{second-order cone program}
\newacro{MPC}{model predictive control}
\newacro{SQP}{sequential quadratic programming}
\newacro{OCP}{optimal control problem}
\newacro{LSTM}{long short-term memory}
\newacro{RNN}{recurrent neural network}
\newacro{GRU}{gated recurrent unit}
\newacro{MSE}{mean-squared-error}
\newcommand{\toast}{TOAST}

\newcommand{\eg}{{e.g.}}
\newcommand{\ie}{{i.e.}}

\newcommand{\reals}{{\mathbf{R}}}

\newcommand{\nf}{n_{\mathrm{f}}}

\definecolor{wheat}{rgb}{0.96,0.87,0.70}

\definecolor{LightCyan}{rgb}{0.88,1,1}
\pdfoutput=1

\title{Constraint-Informed Learning for Warm Starting Trajectory Optimization}

\author[a,b]{Julia Briden\footnote{Doctoral Student, Department of Aeronautics and Astronautics, Massachusetts Institute of Technology; jbriden@mit.edu. AIAA Student Member (Corresponding Author).}}
\author[a]{Changrak Choi\footnote{Robotics Technologist, Jet Propulsion Laboratory, California Institute of Technology.}}
\author[a]{Kyongsik Yun\footnote{Technologist, Jet Propulsion Laboratory, California Institute of Technology.}}
\author[b]{Richard Linares\footnote{Rockwell International Career Development Professor and Associate Professor, Department of Aeronautics and Astronautics, Massachusetts Institute of Technology, Senior Member AIAA.}}
\author[a]{Abhishek Cauligi\footnote{Robotics Technologist, Jet Propulsion Laboratory, California Institute of Technology.}}

\affil[a]{NASA Jet Propulsion Laboratory, California Institute of Technology, 4800 Oak Grove Dr, Pasadena, CA 91109 USA}
\affil[b]{Department of Aeronautics and Astronautics, Massachusetts Institute of Technology, 77 Massachusetts Avenue, Cambridge, Massachusetts, 02139 USA}

\begin{document}

\maketitle

\begin{abstract}
Future spacecraft and surface robotic missions require increasingly capable autonomy stacks for exploring challenging and unstructured domains, and trajectory optimization will be a cornerstone of such autonomy stacks. However, the nonlinear optimization solvers required remain too slow for use on relatively resource-constrained flight-grade computers. In this work, we turn towards amortized optimization, a learning-based technique for accelerating optimization run times, and present~\toast{}: Trajectory Optimization with Merit Function Warm Starts. Offline, using data collected from a simulation, we train a neural network to learn a mapping to the full primal and dual solutions given the problem parameters. Crucially, we build upon recent results from decision-focused learning and present a set of decision-focused loss functions using the notion of merit functions for optimization problems. We show that training networks with such constraint-informed losses can better encode the structure of the trajectory optimization problem and jointly learn to reconstruct the primal-dual solution while yielding improved constraint satisfaction. Through numerical experiments on a Lunar rover problem and a 3-degrees-of-freedom Mars powered descent guidance problem, we demonstrate that~\toast{} outperforms benchmark approaches in terms of both computation times and network prediction constraint satisfaction.
\end{abstract}

\section*{Nomenclature}

{\renewcommand\arraystretch{1.0}
\noindent\begin{longtable*}{@{}l @{\quad=\quad} l@{}}
$N$ & number of discretization intervals \\
$\theta$ & vector of problem parameters, $\theta \in \reals^{n_p}$ \\
$x_t$ & state at time step $t$, $x_t \in \reals^{n_x}$ \\
$u_t$ & control input at time step $t$, $u_t \in \reals^{n_u}$ \\
$x^*$ & optimal state trajectory \\
$u^*$ & optimal control trajectory \\
$\lambda^*$ & optimal dual multipliers \\
$\hat{x}$ & predicted state trajectory \\
$\hat{u}$ & predicted control input \\
$\hat{\lambda}$ & predicted dual multipliers \\
$z = (x, y, \xi, \delta, v, \alpha)^T$ & configuration of the center point of the rear axle of the lunar rover in a fixed world frame $(x,y,\xi)$\\
$u = (\delta_{\text{in}}, \alpha_{\text{in}})^T$ & control of the steering angle control input $\delta_{\text{in}}$ and acceleration control input $\alpha_{\text{in}}$ \\
$g_t$ & stage cost function at time step $t$ \\
$\psi_t$ & state transition function at time step $t$ \\
$f_{t,i}$ & inequality constraint function at time step $t$ for constraint $i$ \\
$\Theta$ & admissible set of parameters, $\Theta \subseteq \reals^{n_p}$ \\
$\phi$ & parameters of neural network \\
$\mathcal{D}$ & training dataset \\
$\mathcal{L}$ & Lagrangian of the optimization problem \\
$\beta_{\mathcal{L}}$ & weighting parameter for the Lagrangian in the merit function \\
$\beta_{\mathcal{\nabla L}}$ & weighting parameter for the Lagrangian gradient in the merit function \\
$\mathcal{I}$ & set of inequality constraints \\
$\mathcal{E}$ & set of equality constraints
\end{longtable*}}

\section{Introduction}
\lettrine{S}{urface} rovers have a rich history of use in planetary exploration, and onboard autonomy has played a critical role in enabling new scientific discoveries.
For example, Mars rover missions such as NASA's \textit{Curiosity} and \textit{Perseverance} have driven tens of kilometers through their mission lifetimes, and autonomous driving capabilities such as~\ac{ENav} have played a crucial part in enabling the rovers to carry out valuable \textit{in situ} measurements and scientific operations~\citep{RankinMaimoneEtAl2020,VermaMaimoneEtAl2023}. However, current rover missions operate at relatively low driving speeds, allowing~\ac{ENav} to utilize a simple search-based approach that outputs geometric paths without considering the full system's high-fidelity dynamics, state, and actuator constraints. Future missions that call for operating at significantly faster speeds will require planners that include a trajectory optimization layer and allow the rover to plan trajectories that fully satisfy kinodynamic constraints.

However, Mars rover operations depend heavily on ground-in-the-loop involvement, particularly for navigating difficult or cluttered terrain. This reliance on ground operations limits the distances that rovers can autonomously travel. For instance, the most recent Decadal Survey~\citep{NASEM2022} highlighted the Endurance Lunar rover mission, a long-range surface rover mission that requires driving several kilometers a day~\citep{KeaneTikooEtAl2022}. Consequently, future rover mission concepts that involve driving significantly further distances than \emph{Curiosity} or \emph{Perseverance} require (1) greater onboard autonomy to minimize ground-in-the-loop interventions while (2) operating at significantly faster speeds. Compared to the \emph{Curiosity} rover, \emph{Perseverance} is equipped with more autonomy for planning and utilizes~\ac{ENav}, a search-based planning approach that generates paths for the rover to follow~\citep{ToupetDelSestoEtAl2020,DaftryAbcouwerEtAl2022}. Follow-on work has further demonstrated the promise of learning-based approaches to enable faster planning~\cite{DaftryAbcouwerEtAl2022}.

Powered descent guidance for planetary landing is known to be particularly challenging for onboard computation; fuel optimal diverts for a general set of state and control constraints require the use of custom solvers to achieve sub-second-level predictions~\cite{Dueri2017, Elango2022, kamath2023customized, kamath2022realtime}. Increasingly complex missions, including Artemis II and Mars Sample Return, will require long-horizon trajectory optimization problems to be solvable online.

Despite the significant advancements in onboard numerical optimization solvers, flight-grade computers remain significantly resource-constrained, lacking the computational power required to solve trajectory optimization problems at the speeds required for real-time operation~\citep{ErenPrachEtAl2017}. Recently, amortized optimization has emerged as a promising solution, leveraging data-driven methods to learn problem-solution mappings, significantly reducing the runtimes required to solve nonlinear optimization problems online~\citep{Amos2023}. However, fully amortized approaches often function as black-box models, learning direct mappings from problem parameters to outcomes without explicitly incorporating the optimization problem's objective or constraints. A field that has been working to overcome these limitations is semi-amortized methods~\cite {Amos2023, KimWisemanEtAl2018, MarinoYueEtAl2018}. These methods integrate the optimization problem's objectives and constraints into their models, preserving awareness of physical and safety constraints. Similar advancements include methods that adjust loss function corrective terms to ensure compliance with equality and inequality constraints~\cite{DontiRolnickEtAl2021}. While these new methods make significant progress in incorporating problem-specific information into the learning process, previous work has been problem-specific, including only variational autoencoder architectures, and cannot be easily applied to general neural network architectures and constrained optimization problems. Further, while semi-amortized optimization-based methods incorporate problem-specific information into the learning process, they are often much slower than amortized optimization, requiring multiple expensive backpropagation steps \cite{KimWisemanEtAl2018}.

In this work, we seek to bridge these gaps and develop a semi-amortized optimization approach for efficiently solving trajectory optimization problems on resource-constrained hardware. Including examples of a planetary rover mission and a spacecraft powered descent application. The new algorithm was developed with the following desiderata in mind:
\begin{enumerate}
    \item {\em Performant: } The controller should yield high-quality or near-optimal solutions with respect to some task metric.
    \item {\em Decision-focused: } The semi-amortized optimization approach should be cognizant of the constraints enforced by the controller downstream.
    \item {\em Generalizable: } The solution approach should not be tailored to a specific problem formulation and apply to a host of future missions requiring on-board trajectory optimization.
\end{enumerate}

The advances in warm-starting, or initial guess generation, provided in this work have the capability to significantly progress the state-of-the-art for future autonomous surface rover missions and powered descent guidance.

\begin{figure}[!t]
\centering
\includegraphics[width=1.0\columnwidth,trim={2cm 14cm 2cm 13cm},clip]{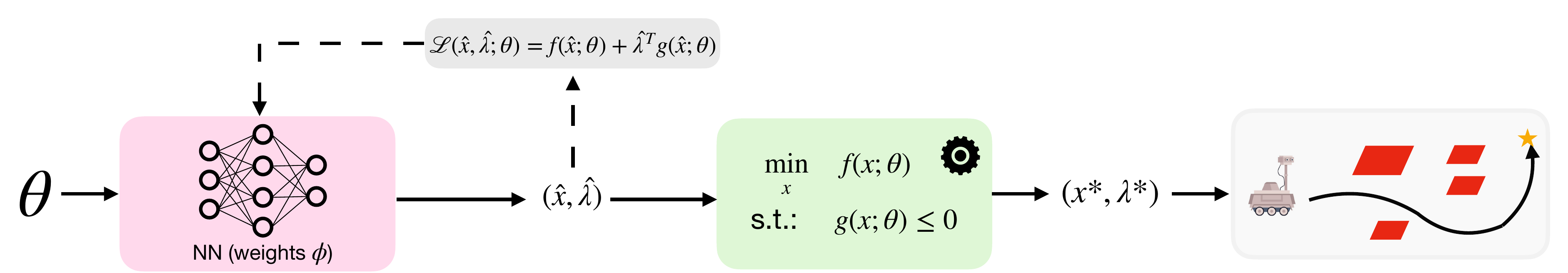}
\caption{Schematic of the \toast{} approach: learning warm starts for optimization using task-relevant merit functions. Warm starts (green) are predicted by the neural network (pink), which is trained offline with a Lagrangian-based merit function.}
\label{fig:alg_outline}
\end{figure}

\subsection{Related Work}

In recent years, there has been a flurry of work on applying data-driven and amortized optimization-based techniques, or learning to predict the solutions to similar instances of the same problem, for accelerating solution times for optimization problems~\citep{KotaryFiorettoEtAl2021,CauligiCulbertsonEtAl2022}. These techniques approach the problem of accelerating solution times for numerical optimization-based control through the lens of parametric programming, a technique to build a function $f : \theta \rightarrow x^*$ that maps the parameters, or context $\theta \subseteq \Theta$, of an optimization problem, $\mathcal{P}(\theta)$, to its solution, $x^*  (\theta) \in \reals^{n_x}$~\citep{DuaKouramasEtAl2008}.
This is accomplished by sampling $\theta$ representative of the problems of interest,  solving for the $x^*$ corresponding to these $\Theta$, and then training an approximation $\hat{f}$ via supervised learning~\citep{Amos2023}.

\cite{DeLaCroixRossiEtAl2024,GhoshTomarEtAl2024}

In contrast to fully-amortized methods, semi-amortized optimized models map the parameters, or context $\theta \subseteq \Theta$, of an optimization problem, $\mathcal{P}(\theta)$, to its solution, $x^*  (\theta) \in \reals^{n_x}$ while accessing the objective function of the optimization problem, often iteratively~\citep{Amos2023}. Previous work has established semi-amortized models for variational inference, allowing for the integration of solvers to improve prediction performance~\citep{KimWisemanEtAl2018, MarinoYueEtAl2018}. These semi-amortized methods involve additional iteration steps over the domain $\mathcal{Y}$ or a latent space $\mathcal{Z}$. Commonly, the optimization procedure is parameterized and integrated into the semi-amortized model $\hat{y}_\theta$, creating a bi-level setting if an outer-level learning problem and inner-level optimization problem~\citep{FinnAbbeelEtAl2017, KimWisemanEtAl2018, AndrychowiczDenilEtAl2016}. Due to the computational efficiency requirements for real-time trajectory generation, this work maintains the idea of infusing the optimization process in the learning process, but we do not merge the two processes; to maximum runtime efficiency, all training of learning-based methods occur offline, where the optimization process informs the learning process, then only the learned initial guess is utilized online to warm-start the optimization solve. Maintaining the underlying feasibility and optimality guarantees of the solver.

Applications of amortized optimization for warm-starting trajectory optimization have shown tremendous promise in robotics. 
The authors in~\citep{ChenWangEtAl2022,SambharyaHallEtAll2022,MorelliHofmannEtAl2024} propose using a neural network to warm start solutions for a~\ac{QP}-based~\ac{MPC} controller. Additional works have studied extensions for quickly solving non-convex optimal control problems online. In~\citep{IchnowskiAvigalEtAl2020}, the authors train a neural network to learn the problem solution mapping for a non-convex robotic grasp optimization problem solved using~\ac{SQP}. The authors in~\citep{BridenGurgaEtAl2024} and~\citep{GuffantiGammelliEtAl2024} train transformer neural networks to learn efficient warm-starts for numerical optimization solvers for trajectory optimization, including applications in powered descent guidance and spacecraft rendezvous. A learned mapping between problem parameters to the set of active tight constraints and final times was shown to reduce solution times for a powered descent guidance problem by more than an order of magnitude~\citep{BridenGurgaEtAl2024}. In spacecraft rendezvous, the transformer-generated warm-starts converge to more fuel-efficient trajectories and higher constraint satisfaction, when compared to convex relaxation benchmarks~\citep{GuffantiGammelliEtAl2024}. Not all approaches overlook the utilization of the learned mappings in relation to the trajectory optimization problem's structure. For example, the work of~\citep{SambharyaStellato2024} demonstrates a methodology that integrates the underlying structure of the optimization problem by unrolling algorithm steps in the Douglas-Rachford (DR) splitting to solve the~\ac{QP}. Further improvements in interior point method-based warm starts for Successive Convex Programming (SCP) include utilizing the solution from the previous SCP iteration while developing an indicator for the degree of problem difference~\citep{MorelliHofmannEtAl2024}. Our work aims to extend these works in semi-amortized optimization by creating a set of generalized merit functions that balance cost minimization with constraint satisfaction. These functions are generalizable to any constrained optimization problem.

\subsection{Statement of Contributions}
In this work, we introduce Trajectory Optimization with Merit Function Warm Starts~(\toast{}), a framework designed to bridge the gap in the aforementioned fields of amortized optimization and nonlinear trajectory optimization. TOAST incorporates two separate phases to warm-start a general set of constrained optimization problems: 1) \textit{Offline Learning}: a neural network is trained to map the problem parameters to the time-varying policy associated with a non-convex trajectory optimization problem and 2) \textit{Online Inference and Solve}: the network and system dynamics are used to predict the full state and control trajectories for a new problem, and this prediction is used to warm start the numerical optimization solver. Rather than learning the fuel control policy and solution trajectory, we emphasize that only the control policy is learned by the neural network. The policy is then propagated through the full system dynamics to enforce the dynamic feasibility of the initial guess. The neural network formulation includes two architecture options: a recurrent neural network (RNN)-based long short-term memory (LSTM) architecture to improve model compactness and enforce the temporal structure of the~\ac{OCP} and a transformer architecture for more complex~\ac{OCP}s. The primary contribution of this work is developing a set of constraint-informed merit functions used for computing loss for the neural network during the offline learning phase. Informed by the Lagrangian and the KKT conditions from optimization theory, this novel set of \emph{decision-focused} loss functions jointly learn to minimize the reconstruction error and the feasibility of the prediction using the merit function associated with the underlying trajectory optimization problem. We show through numerical simulations on a surface rover trajectory planning problem and a powered descent guidance problem that our proposed~\toast{} approach outperforms benchmark amortized optimization approaches with improved constraint satisfaction. As future rover missions incorporate increased onboard decision-making, dynamically feasible trajectory generation will be required to achieve strategic planned trajectories accurately. TOAST significantly improves the computational efficiency of onboard trajectory planning via initial guesses biased toward constraint satisfaction.

\textit{Paper Organization: }
This work is organized as follows: Section \ref{sec:technical_background} reviews the technical background for our approach, including the terminology for the learned solution mapping, optimization theory background, deep learning architectures, and an overview of decision-focused learning. Our technical approach is covered in Section \ref{sec:technical_approach}, which describes the parametric machine learning problem, control input prediction, and dynamics propagation process and introduces our constraint-informed merit function warm start framework: Trajectory Optimization with Merit Function Warm Starts (TOAST). A set of numerical experiments evaluate the TOAST's performance for Lunar rover model predictive control (MPC) and Mars powered descent guidance in Section \ref{sec:application}, including introducing the lunar rover problem and powered descent guidance problem formulations, evaluating loss function sensitivity, and evaluating the accuracy and constraint satisfaction for the TOAST LSTM framework, compared to mean-squared-error (MSE) loss. Finally, takeaways and conclusions for this work are reviewed in Section \ref{sec:conclusion}.

\section{Preliminaries}\label{sec:technical_background}

\subsection{Learning a Solution Map}
Given a vector of problem parameters $\theta\in\reals^{n_p}$, a parametric~\ac{OCP} can be written as
\begin{equation} \label{eq:nlp}
\begin{array}{ll}
\underset{x_{0:N},u_{0:N}}{\textrm{minimize}} \!\!\!& \sum_{t=0}^{N} g_t(x_t,u_t;\theta) \\
\text{subject to}\!\!\!& x_0 = x_\textrm{init}(\theta),\\
& x_{t+1} = \psi_t(x_t,u_t;\theta), \quad t = 0, \dots, N-1,\\
& f_{t,i}(x_t,u_t;\theta) \leq 0, \quad\;\;\;\, t = 0, \dots, N, i = 1,\dots,\nf, \\
\end{array}
\end{equation}
where the state $x_t\in\reals^{n_x}$ and control $u_t\in\reals^{n_u}$ are the continuous decision variables.
Here, the stage cost $g_t(\cdot)$ and terminal cost $g_N(\cdot)$ are assumed to be convex functions, but the dynamical constraints $\phi_t (\cdot)$ and inequality constraints $f_{t,i}(\cdot)$ are assumed smooth but possibly non-convex. The objective function and constraints are functions of the parameter vector $\theta \in \Theta$, where $\Theta \subseteq \reals^{n_p}$ is the admissible set of parameters.

In robotics, the~\ac{OCP} is typically solved in a receding horizon fashion as the controller replans periodically, wherein the problem size typically stays fixed. Still, only the problem parameters $\theta$ vary between repeated optimization calls.
This setting motivates learning function $f$ that maps problem parameters $\theta$ to the optimal solution $x^*$ for the~\ac{OCP}, as the learned mapping can be utilized directly (\eg{}, imitation learning) or as a warm start initialization for the solver.


\subsection{Necessary Conditions for Optimality: KKT Conditions}
 For an inequality-constrained optimization problem, Equation \ref{eq:nlp}, the first order necessary conditions, or Karush-Kuhn-Tucker (KKT) conditions, must hold at a local optimum. If we take the boundary conditions in Equation \ref{eq:nlp}, to be defined inside of the set of equality constraints $\psi$, the KKT conditions are as follows,
    
    \begin{definition}[First Order Necessary Conditions: KKT Conditions]
        
        \[
        \mathcal{L}(z, \lambda) = \sum_{t=0}^{N} g_t(z_t) - \sum_{i \in \mathcal{I}} \lambda_i f_{t,i}(z_t) - \sum_{i \in \mathcal{E}} \lambda_i \psi_{t,i}(z_t), \; t = 0, \dots, N,
        \]
        
        \begin{enumerate}
            \item $\nabla_x \mathcal{L}(z^*, \lambda^*) = 0$,
            \item $\psi(z^*) = 0, \quad i \in \mathcal{E}$,
            \item $f_i(z^*) \geq 0, \quad i \in \mathcal{I}$,
            \item $\lambda_i^* \geq 0, \quad i \in \mathcal{I}$,
            \item $\lambda_i^* f_i(x^*) = 0 \; \text{and} \; \lambda_j^* \psi_j(z^*) = 0, \quad i \in \mathcal{I}, \; j \in \mathcal{E}$.
        \end{enumerate}
    
    \end{definition}

The sets $\mathcal{I}$ and $\mathcal{E}$ are the sets of indices for the inequality and equality constraints and $z$ represents the set of decision variables $z = (x, u)$ for the \ac{OCP}. Condition 1 denotes dual feasibility, conditions 2-3 include primal feasibility, and conditions 4-5 define complementary slackness. The Lagrangian $\mathcal{L}(z, \lambda)$, which includes the cost function $\sum_{t=0}^{N} g_t(z_t)$ and constraints $f_t$ and $\psi_t$, is equivalent to only the cost $\sum_{t=0}^{N} g_t(z_t)$ when the KKT conditions are met (by condition 5). Further, the gradient of the Lagrangian is zero when evaluated at a KKT point (by condition 1). Not only does the Lagrangian balance cost and constraint satisfaction, but minimization of the Lagrangian naturally meets the KKT conditions, necessary for optimality. This observation informs our choice of the Lagrangian and Lagrangian gradient for our set of constraint-informed merit functions in Section \ref{sec: merit functions for warm starts}.

\subsection{Recurrent Neural Network Architectures}
Unlike feedforward neural networks, \ac{RNN} architectures allow the use of previously gathered information to inform the current decision. Within the context of amortized optimization, recent works have shown how the inherently Markovian structure of the inference procedures used by~\ac{RNN} architectures such as~\ac{LSTM} and~\ac{GRU} can be used for tackling nonlinear trajectory optimization problems~\citep{SabolYunEtAl2022,CauligiChakrabartyEtAl2022}. In this work, we extend these~\ac{LSTM} frameworks to study the problem of learning warm starts for long-horizon nonlinear trajectory optimization problems in the context of decision-focused learning.

\subsection{Transformer Neural Network Architectures}

Transformer neural networks (NNs) have shown improved performance in training long-horizon time series data, mitigating the vanishing or exploding gradient problem and allowing for parallelization during training \cite{Vaswani2017}. Multi-head attention mechanisms enable different parts of the input to garner attention while processing entire sequences in parallel. This work utilizes a transformer NN architecture to predict the control input and propagate the trajectory for the 3 DoF powered descent guidance problem, training the TOAST transformer architecture with the constraint-informed merit functions discussed in the following section.


\subsection{Decision-Focused Learning}
Amortized optimization has recently gained attention for its potential to improve the computational efficiency of optimal control problems. However, fully amortized models use loss functions that overlook the integration of the trajectory optimization problem’s intrinsic structure, as these models generally learn direct mappings from input parameters to solutions without explicit regard for the optimization’s constraints and objectives. In contrast, semi-amortized methods actively incorporate these elements, utilizing detailed information about the optimization problem to inform the loss function. In this work, we draw inspiration from decision-focused learning, which integrates neural network training directly with operational decision-making \citep{WilderDilkinaEtAl2019,MandiKotaryEtAl2023}. Unlike traditional learning-based approaches that utilize a standard catalog of loss functions, decision-focused learning customizes these functions to reflect the specific parameters and constraints of the optimization problems, aligning closely with the goals of semi-amortized approaches. This approach ensures that the learning process supports the practical deployment of models in decision-making scenarios, which is particularly relevant in control tasks where adherence to physical and safety constraints is crucial. While our approach is inspired by decision-focused learning, decision-focused learning involves integrating the predictions of uncertain quantities into the decision-making process, which differs from the deterministic nature of the direct optimization problems we address.

To extend such semi-amortized approaches, we turn to the solution techniques used for solving constrained optimization to formalize the concept of decision-focused losses and consider \textit{merit functions}, which are a scalar-valued function of problem variables that indicate whether a new iterate is better or worse than the current iterate, with the goal of minimizing a given function~\citep{NocedalWright2006}.
Although a candidate merit function for unconstrained optimization problems is the objective function, a merit function for a constrained optimization problem must balance minimizing the cost function with a measure of constraint violation.
For example, an admissible merit function for~\eqref{eq:nlp} is the penalty function of the form $\phi(x,\mu) = f(x) + \mu \sum_{i \in \mathcal{E}} |c_i(x)| + \mu \sum_{i \in \mathcal{I}} [ c_i(x) ]^-$, where $[ c_i(x) ]^- = \max\{0, -c_i(x)\}$ (we refer the reader to~\citep{NocedalWright2006} for a more exhaustive discussion and examples of merit functions). \toast{} generalizes this definition of a merit function to develop a set of decision-focused loss function formulations to facilitate effective warm-starts for online \ac{MPC}.

\section{Technical Approach}\label{sec:technical_approach}

In this work, we seek to learn a solution mapping $f (\theta)$ that maps problem parameters $\theta$ to the optimizer $z^* = (x^*, u^*)$, where $x$ are the state trajectories and $u$ are the control inputs. This can be accomplished by approximating the solution map $f (\theta)$ using a deep neural network $f_\phi (\theta)$, wherein $\phi$ are the neural network parameters to be learned. By formulating this problem as a parametric machine learning problem, a dataset $\mathcal{D} = ((\theta_i, z_i))_{i=1}^n$, a parameterized function class $f_\phi$, and a loss function $L(z,f_\phi (\theta))$ are user-specified and the goal of the learning framework is to compute $\phi$ such that the expected risk on unseen data is minimized, $\min_\phi \mathbb{E}[L(z, f_\phi(\theta))]$. Note that the minimum expected risk over unseen data cannot be computed since we cannot access all unseen data. Assuming the training set is a sufficient representation of the unseen data, $\min_{\phi} L(z,f_\phi (\theta))$, the empirical risk (training loss) will well-represent the expected risk (test loss).

If $\hat{z}$ denotes the full primal solution prediction $(\hat{x}, \hat{u})$, then the ``vanilla'' approach to accomplish this would be to simply model $f_\phi (\cdot)$ as a regressor and output a prediction $\hat{z}$ for the full primal solution, \ie{}, $f_\phi (\theta) = \hat{z}$.
However, this approach has the shortcoming that predicting the full primal solution $\hat{z} \in \reals^{N(n_x+n_u)}$ can be challenging to supervise due to the large output dimensionality and current approaches do indeed scale poorly with increasing state dimension and time horizon~\citep{ChenWangEtAl2022, ZhangBujarbaruahEtAl2019}.

\begin{figure}[!t]
\centering
\includegraphics[width=.7\linewidth]{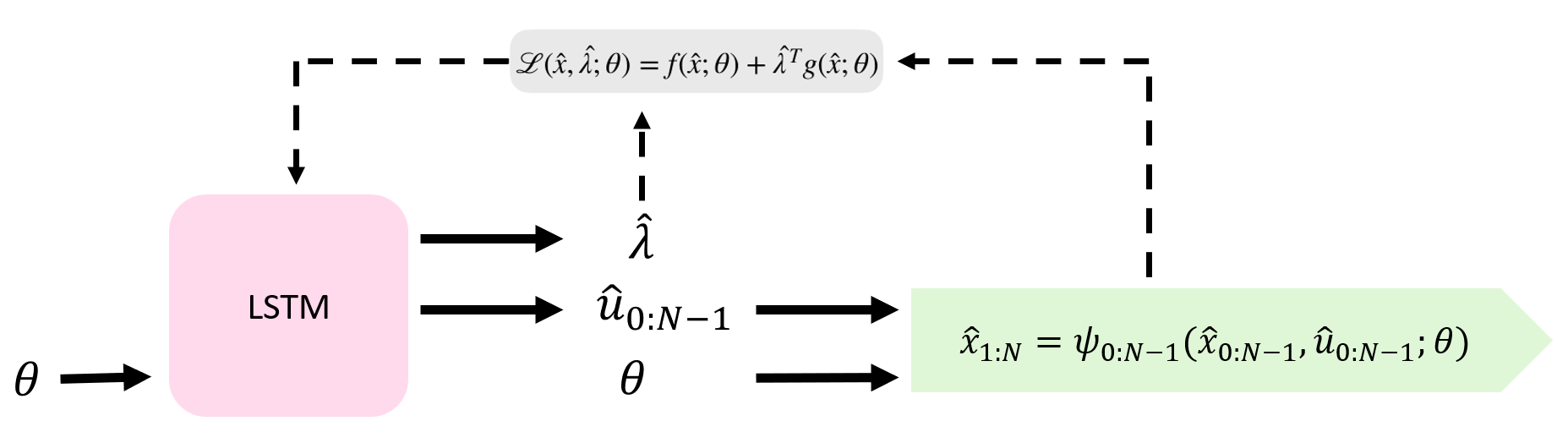}
\caption{Only the control policy and dual variables are predicted using the LSTM, and the state variables are recovered by propagating the controls through the system dynamics. During the learning process, the Lagrangian-based merit function uses the decision variables from the propagated dynamics to evaluate the chosen loss function.}
\label{fig:propagate}
\end{figure}

Rather than learning the full mapping from $\theta$ to $\hat{z}$, we instead learn a time-varying policy $\pi_\tau (\cdot)$, \ie{}, $$\{ u_\tau \}_{\tau=0}^{N} = \{ \pi_\tau (\theta) \}_{\tau=0}^{N}$$ (see Figure \ref{fig:propagate}). Given $x_0 = x_\textrm{init}(\theta)$, the state prediction $\hat{x}$ is recovered by simply forward propagating the dynamics:
\begin{equation*}
\hat{x}_{t+1} = \psi_t(\hat{x}_t, \hat{u}_t; \theta), \quad t = 0, \dots, N-1.\\
\end{equation*}

The advantages of using an~\ac{RNN} to learn the time-varying policy include:
\begin{enumerate}
    \item Using a more compact neural network model for $f_\phi (\cdot)$ that has output dimension in $\reals^{n_u}$ and rather than in $\reals^{N(n_x+n_u)}$.
    \item Incorporating the temporal structure of the problem via the~\ac{RNN} in predicting decision variables.
\end{enumerate}

We note that our proposed approach is closely related to the area of solving model-based trajectory optimization for imitation learning, an area of research that has extensive heritage~\citep{ReskeCariusEtAl2021,TagliabueKimEtAl2022,CauligiChakrabartyEtAl2022}. However, we eschew the imitation learning terminology since our proposed approach still relies on running numerical optimization online, thereby preserving any guarantees of the underlying solver. We also note connections to shooting methods~\citep{Betts1998,Kelly2017}, wherein the number of decision variables in a trajectory optimization problem is reduced by only optimizing over the controls $\{ u_t \}_{t=0}^N$. As is well known, however, shooting methods often struggle to find solutions for problems with challenging state constraints, but~\toast{} addresses this challenge by jointly predicting the dual variables for improved constraint satisfaction predictions.

The next section discusses our proposed approach to generating predictions that better satisfy system constraints.

\subsection{Merit Functions for Warm Starts}\label{sec: merit functions for warm starts}

\begin{figure}[ht]
        \centering
        \includegraphics[width=.8\linewidth]{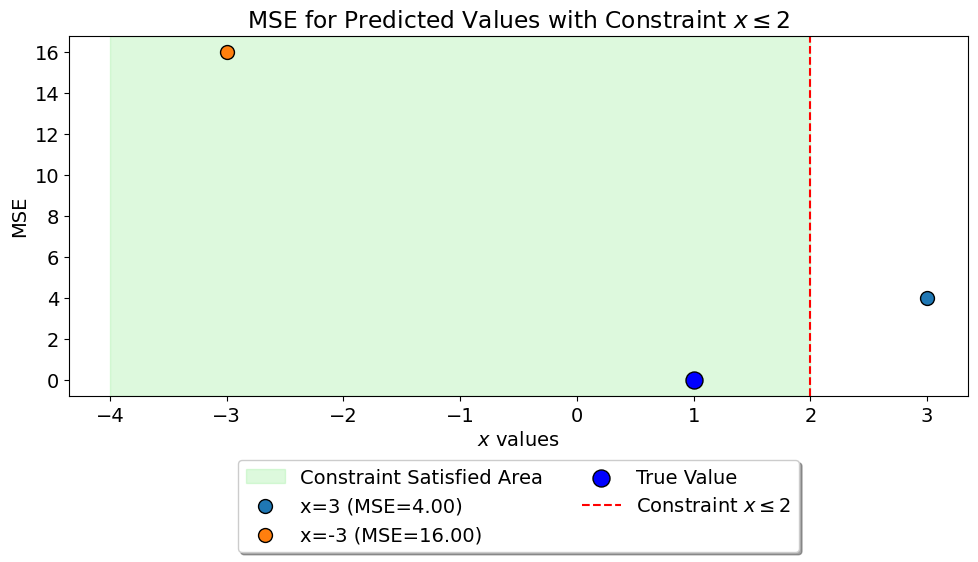}
        \caption{Mean-squared-error (MSE) for predicting $x = 1$ with constraint $x \leq 2$.}
  \label{fig:MSE}
  \end{figure}

To motivate the need for merit function-based learning, we consider the following simple optimization problem:

\begin{align*}
\text{minimize:} \quad & \text{MSE} = \frac{1}{N} \sum_{i=1}^{N} (x^*_i - \hat{x}_i)^2 \\
\text{subject to:} \quad & x \leq 2,
\end{align*}

where $x^*$ is the optimal value of the decision variable and $\hat{x}$ is the prediction, for $N$-dimensional $x$. Figure \ref{fig:MSE} demonstrates an example of the MSE for two separate predictions when the optimal value is $x^* = 1$. While the prediction $x = 3$ has a lower MSE, the prediction $x = -3$ is constraint satisfying. For safety-critical robotic applications, where constraint satisfaction is required for mission success, the constraint-satisfying prediction is often valued over cost minimization. In this work, we develop task-relevant merit functions that balance this trade-off in constraint satisfaction and cost minimization.

Let $\hat{z}(\theta) =(\hat{x}, \hat{u})$ be the initial prediction for the continuous decision variables of~\eqref{eq:nlp} by a neural network model $f_\phi (\theta)$ for problem parameters $\theta$.
The standard training loss for updating the parameters $\phi$ of the model would be the~\ac{MSE} loss function $L_{\text{MSE}} = \min_\phi \frac{1}{|\mathcal{D}|} \sum_{i=1}^{|\mathcal{D}|} \| \hat{z}_i(\theta) - z_i^*\|_2^2$, where $\mathcal{D}$ is the training set of tuples $\{ (\theta_i, z_i^*) \}_{i=1}^{|\mathcal{D}|}$ constructed by solving~\eqref{eq:nlp} to optimality.

When comparing~\ac{MSE} loss to the set of~\toast{} merit functions, detailed below, we will benchmark our results against both~\ac{MSE} and Primal~\ac{MSE}. ~\ac{MSE} loss computes the mean-squared-error of the state, control input, and dual variables, and Primal~\ac{MSE} loss computes the mean-squared-error of the state and control decision variables only.

This work uses merit functions as decision-focused loss functions to supervise a problem-solution map for constrained optimization problems. Instead of using the standard~\ac{MSE} loss, we seek to generate better solution predictions $\hat{z}$ that allow for faster online convergence by explicitly penalizing system constraint violations. To accomplish this, we propose the following set of merit functions:

\begin{enumerate}
\item \textbf{Lagrangian Loss} \begin{equation}
L_{\mathcal{L}} = \min_\phi \frac{1}{|\mathcal{D}|} \sum_{i=1}^{|\mathcal{D}|} (\mathcal{L}(z^*; \theta) - \mathcal{L}(\hat{z}; \theta))^2, \label{eq:loss_lag_diff}
\end{equation}

where $\mathcal{L}(\hat{z}; \theta) = f(\hat{z}; \theta) + \hat{\lambda}^T g(\hat{z}; \theta)$ is the Lagrangian associated with~\eqref{eq:nlp} evaluated at $\hat{z}$, $f$ is the cost function for the optimization problem, $\hat{\lambda}$ are the dual multipliers, and $g$ is a vector of constraints.
Equality constraints are not shown in this formulation since all equality constraints are separated into two inequality constraints in the numerical examples.

\item \textbf{Lagrangian with Gradient Loss}
\begin{equation}
L_{\nabla \mathcal{L}} = \min_\phi \frac{\beta_{\mathcal{L}}}{|\mathcal{D}|} \sum_{i=1}^{|\mathcal{D}|} (\mathcal{L}(z^*; \theta) - \mathcal{L}(\hat{z}; \theta))^2 + \beta_{\mathcal{\nabla L}}(\nabla_z \mathcal{L}(\hat{z}_i; \theta))^2. \label{eq:loss_lag_diff_kkt}
\end{equation}

This loss function follows from adding the KKT conditions' stationarity condition for optimization problems~\cite{BoydVandenberghe2004,NocedalWright2006}.
Instead of using the stationarity condition alone, it was instead added to Lagrangian loss since learning with $(\nabla_z \mathcal{L}(\hat{z}_i; \theta))^2$ only often led to maximization, instead of minimization of the loss during learning. The parameters $\beta_{\mathcal{L}}$ and $\beta_{\mathcal{\nabla L}}$ are adjustable multipliers for scaling each quantity. For the numerical experiments in this work, $\beta_{\mathcal{L}} = 0.1$ and $\beta_{\mathcal{\nabla L}} = 0.01$.

\item \textbf{Lagrangian MSE Loss} \begin{equation}
L_{\mathcal{L} \text{ MSE}} = \min_\phi \frac{\beta_{\mathcal{L}}}{|\mathcal{D}|} \sum_{i=1}^{|\mathcal{D}|} (\mathcal{L}(z^*; \theta) - \mathcal{L}(\hat{z}; \theta))^2 + \frac{1}{|\mathcal{D}|} \sum_{i=1}^{|\mathcal{D}|} \| \hat{z}_i(\theta) - z_i^*\|_2^2. \label{eq:loss_lag_diff_mse}
\end{equation}

Lagrangian MSE loss is motivated by the regularization of the MSE loss. Consider the estimate $\mathbb{E}[(\hat{z}_i(\theta) - z_i^*)^2]$ which can be decomposed into the sum of $\text{Var}[\hat{z}_i(\theta)] + (\mathbb{E}[\hat{z}_i(\theta)-z_i^*])^2$, the sum of the variance of the predictions and the squared bias of the predictions vs. targets. Given that with an unbiased estimator $\mathbb{E}[\hat{z}] = z$, a high variance could result in a large error. When the estimates are outputs from a NN, we can include a bias term in the loss function; in our case, we have the Lagrangian $\frac{1}{|\mathcal{D}|} \sum_{i=1}^{|\mathcal{D}|} (\mathcal{L}(z^*; \theta) - \mathcal{L}(\hat{z}; \theta))^2$, which biases MSE loss towards constraint satisfaction. Similar to the use of ridge regression in the loss function to reduce overfitting.
\end{enumerate}

\section{Numerical Experiments}\label{sec:numerical_experiments}
In this section, we validate~\toast{} in numerical experiments and focus on the surface rover trajectory optimization problem.
The neural network architectures were implemented using the~\verb|PyTorch| machine learning library~\citep{PaszkeGrossEtAl2017} with the ADAM optimizer~\citep{KingmaBa2015} for training. The optimization problems are modeled using CasADi~\citep{AnderssonGillisEtAl2019} and solved using the IPOPT sequential quadratic programming library~\citep{WachterBiegler2006}.

To benchmark our proposed approach, we compare our decision-focused merit function against vanilla~\ac{MSE} and Primal~\ac{MSE} loss functions for training LSTM NN architectures, as well as the vanilla and collision-penalizing LSTM and feedforward architectures in~\citep{SabolYunEtAl2022}. Table \ref{tab:algorithm_settings} shows the settings used for the lunar rover benchmark problem, and Table 
\ref{tab:algorithm_settings_pdg} shows the settings used for the powered descent guidance problem.

    \begin{table}
        \caption{\label{tab:algorithm_settings} Lunar Rover Settings}
        \centering
        \begin{tabular}{ll}
            \hline
            \textbf{Parameter} & \textbf{Value} \\
            Discretization size ($N$) & 61 \\
            Number of obstacles ($n_{\text{obs}}$) & 5 \\
            Final Time ($T_f$) & 30 s \\
            Wheelbase of the vehicle ($L$) & 2.7 m \\
            Radius ($r_{\text{dist}}$) & 0.075 m \\
            Min steering wheel angle ($\delta_{\min}$) & -90 deg \\
            Max steering wheel angle ($\delta_{\max}$) & 90 deg \\
            Min velocity ($v_{\min}$) & 0 m/s \\
            Max velocity ($v_{\max}$) & 0.2778 m/s \\
            Min acceleration ($\alpha_{\min}$) & -0.3 m/s$^2$ \\
            Max acceleration ($\alpha_{\max}$) & 0.3 m/s$^2$ \\
            Jerk weight ($\omega_{\text{jerk}}$) & 0.5 \\
            Acceleration weight ($\omega_{\text{acceleration}}$) & 1 \\
            \hline
        \end{tabular}
    \end{table}

    \begin{table}
        \caption{\label{tab:algorithm_settings_pdg} Powered Descent Guidance Settings}
        \centering
        \begin{tabular}{ll}
            \hline
            \textbf{Parameter} & \textbf{Value} \\
            Discretization size ($N$) & 21 \\
            Final Time ($T_f$) & 30 s \\
            Gravitational Acceleration ($g$) & 3.72076 m/s$^2$ \\
            Specific Impulse (Isp) & 225 s \\
            Dry Mass ($m_\text{dry}$) & 2200 kg \\
            Minimum Thrust ($\rho_{\min}$) & 18 kN \\
            Maximum Thrust ($\rho_{\max}$) & 48 kN \\
            Max velocity ($v_{\max}$) & 500 m/s \\
            \hline
        \end{tabular}
    \end{table}

To implement the decision-focused loss functions~\eqref{eq:loss_lag_diff}-~\eqref{eq:loss_lag_diff_mse}, the Lagrangian and Lagrangian gradient functions were automatically defined using CasADi's symbolic framework and integrated into~\verb|PyTorch| via AutoGrad classes. When training with dual variables, since~\verb|PyTorch| often returns very large dual variable values (on the order of $1e^6$), clipping is used to prevent instabilities or exploding gradients~\citep{HaeserHinderEtAl2021}. Clipping was chosen instead of an alternative data standardization process since it maintains the physical information of whether a constraint is active or inactive. IPOPT was chosen as the solver since both primal and dual guesses can be provided to the solver, and it can be applied to nonconvex optimization problems.
In practice, we found that providing warm start initializations for the dual variables alone did not affect the solve time for IPOPT. The \toast{} architecture has three layers and 128 neurons.

\subsection{Surface Rover Problem}~\label{subsec:surface_rover_mpc}

We model the dynamics of a lunar rover~\ac{OCP} using the bicycle kinematics model given in Equation 
\ref{eq:state_defn} ~\citep{LiuPadenEtAl2018}:
\begin{equation}
\begin{split}
\dot{z} = \begin{pmatrix}
v \cos\theta\\
v \sin\theta\\
\frac{v}{L} \tan \delta\\
-\lambda_1 \delta + \lambda_1 \delta_\mathrm{in}\\
\alpha \\
-\lambda_2 \alpha + \lambda_2 \alpha_\mathrm{in}\\
\end{pmatrix},
\end{split}
\quad\quad
\begin{split}
z = (x,y,\xi,\delta,v,\alpha)^T,\\
u = (\delta_\mathrm{in}, \alpha_\mathrm{in})^T.
\end{split}
\label{eq:state_defn}
\end{equation}
In this model, the state $z \in \reals^6$ consists of the configuration of the center point of the rear axle of the vehicle in a fixed world frame $(x,y,\xi)$, the steering-wheel angle $\delta$, and the longitudinal speed and acceleration $v$ and $\alpha$, respectively~\citep{LiuPadenEtAl2018}. The control $u\in\reals^2$ consists of the steering angle control input $\delta_\mathrm{in}$ and acceleration control input $\alpha_\mathrm{in}$. The discrete-time update equation is approximated using a backward Euler rule.

The MPC formulation of the lunar rover OCP is then given by:
\begin{equation} \label{eq:lunar rover ocp}
\begin{array}{ll}
\underset{z_{0:N},u_{0:N}}{\textrm{minimize}} \!\!\!& \sum_{t=0}^{N-1} \omega_{\text{jerk}} (\delta_{t+1} - \delta_t)^2 + \sum_{t=0}^N \omega_{\text{acc}} \alpha_t^2 \\
\text{subject to}\!\!\!& z_0 = z_\textrm{init}(\theta),\\
& z_{t+1} \leq \psi_t(z_t,u_t;\theta), \quad t = 0, \dots, N-1,\\
& z_{t+1} \geq \psi_t(z_t,u_t;\theta), \quad t = 0, \dots, N-1,\\
& z_{\min} \leq z_i \leq z_{\max}, \quad\;\;\;\, t = 0, \dots, N, \\
& u_{\min} \leq u_i \leq u_{\max}, \quad\;\;\;\, t = 0, \dots, N-1, \\
& r_{\text{dist}}^2 - [(x_t - r_{\text{obs}_i, x})^2 + (y_t - r_{\text{obs}_i, y})^2] \leq 0, \quad\;\;\;\, t = 0, \dots, N-1, i = 1,\dots,n_{\text{obs}}, \\
\end{array}
\end{equation}
where the cost function minimizes jerk and acceleration over the trajectory and $\omega_{\text{jerk}}$ and $\omega_{\text{acc}}$ are scaling factors.
The dynamics are decomposed into two inequality constraints to allow for only non-negative duality multipliers.
In addition to upper and lower bound constraints for the state and control inputs, collision avoidance constraints are formulated for every obstacle.
The parameters used in this model were drawn from the Endurance-A Lunar rover mission concept studied in the most recent Planetary Decadal survey~\citep{NASEM2022,KeaneTikooEtAl2022}.

\subsection{Loss Function Sensitivity}\label{sec:sensitivity}

\begin{figure}[ht]
        \centering
        \includegraphics[width=\linewidth]{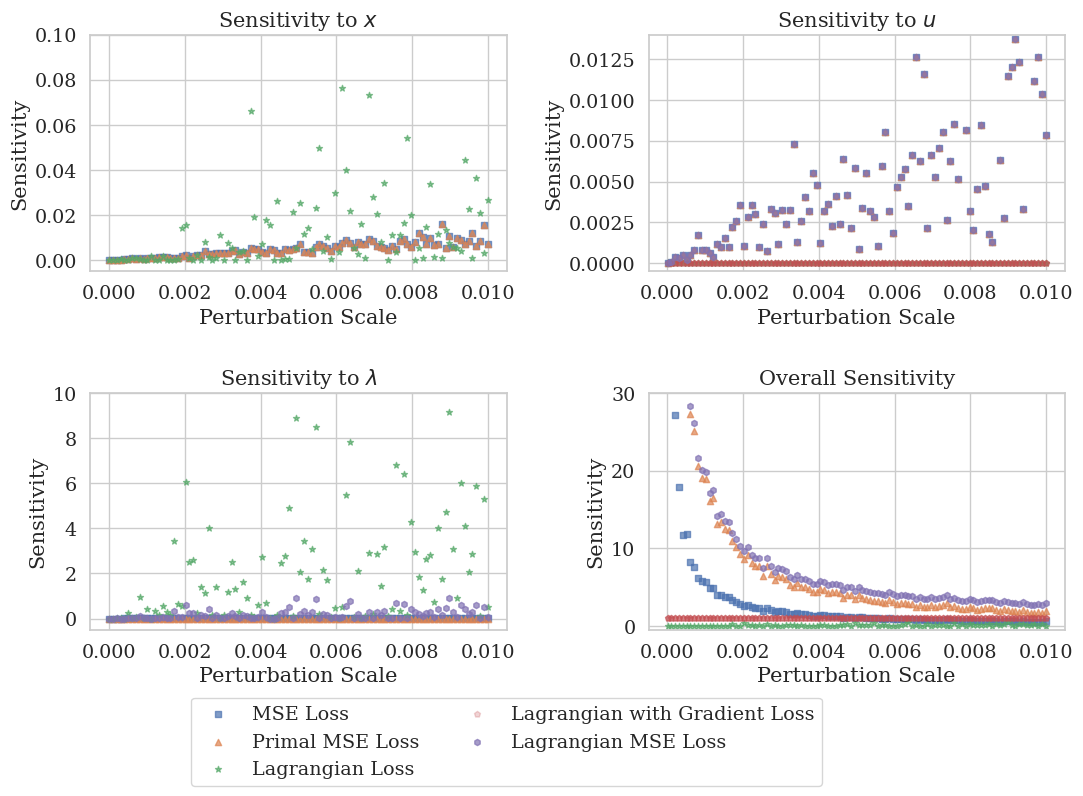}
        \caption{Sensitivity analysis of loss functions: Lagrangian loss displays stable sensitivity; MSE variations decrease across perturbation scales. Integrated Lagrangian MSE loss combines sensitivities to $u$, $x$, and $\lambda$.}
  \label{fig:sensitivity}
  \end{figure}

To understand the loss landscape for the merit functions in Eqns.~\eqref{eq:loss_lag_diff}-~\eqref{eq:loss_lag_diff_mse}, a preliminary sensitivity analysis was conducted. A simple solution to the optimal control problem,~\eqref{eq:lunar rover ocp}, was computed with $N = 4$ discretization nodes and two randomly distributed obstacles. Using the resulting $(x^*, u^*, \lambda^*)$ and clipping $\lambda^* \leq 1$, one hundred perturbations were created equidistantly between the values of $1e^{-6}$ and $1e^{-2}$ to slightly alter the decision variable values at the optimal solution. The sensitivity is then defined as $\frac{L}{\delta}$, where $L$ is the evaluated loss function and $\delta$ is the perturbation scale. The overall sensitivity computations include a combined norm term in the denominator, the sum of the norms of the differences between perturbed solutions. Figure~\ref{fig:sensitivity} shows the sensitivity of each loss function to perturbations in the state $x$, control input $u$, and dual variables $\lambda$.

As is expected, we see that the MSE and Primal MSE loss, where~\ac{MSE} loss computes the mean-squared-error of the state, control input, and dual variables, and Primal~\ac{MSE} loss computes the mean-squared-error of the state and control decision variables only, demonstrated the same sensitivity distribution over $x$ and $u$. Including dual variables $\lambda$ in the MSE prediction likely reduced overall sensitivity because the included dual variables attain values less than or equal to one. For the Lagrangian loss, the largest degree of sensitivity was to the dual variables, and, in contrast to MSE and Primal MSE, the Lagrangian loss was not sensitive to changes in the control.

Overall, the Lagrangian loss on this perturbation scale forms a hyperplane with a slight amount of noise as the perturbation scale increases. When the Lagrangian gradient is added in, and the multipliers $\beta_{\mathcal{L}}$ and $\beta_{\mathcal{\nabla L}}$ are included in the loss, sensitivity is extremely low for all decision variables. Finally, Lagrangian MSE appears to blend the sensitivity to $u$ of the MSE and Primal MSE losses and the sensitivity to $x$ and $\lambda$ from the Lagrangian. The Lagrangian MSE loss function is the only loss function that maintains the approximately monotonically decreasing shape of the MSE loss for overall sensitivity. High sensitivity near the optimal solution for MSE, Primal MSE, and Lagrangian MSE losses could accelerate convergence to minima during training; the areas close to the optimal solution are more responsive to adjustments in the training process, potentially leading to more efficient learning. The low overall sensitivity of the Lagrangian and Lagrangian with Gradient merit functions to small perturbations may be advantageous for training or testing on adversarial examples.

The authors in~\citep{SzegedyZarembaEtAl2014} observe that small adversarial perturbations on input images can change the NN’s prediction. Therefore, since Lagrangian-based merit functions are less susceptible to minor input perturbations, they may be more robust to adversarial examples. Future work will further explore decision-focused learning in adversarial settings.

\subsection{Application: Lunar Rover MPC}\label{sec:application}

\begin{figure}[ht]
        \centering
        \includegraphics[width=.5\linewidth]{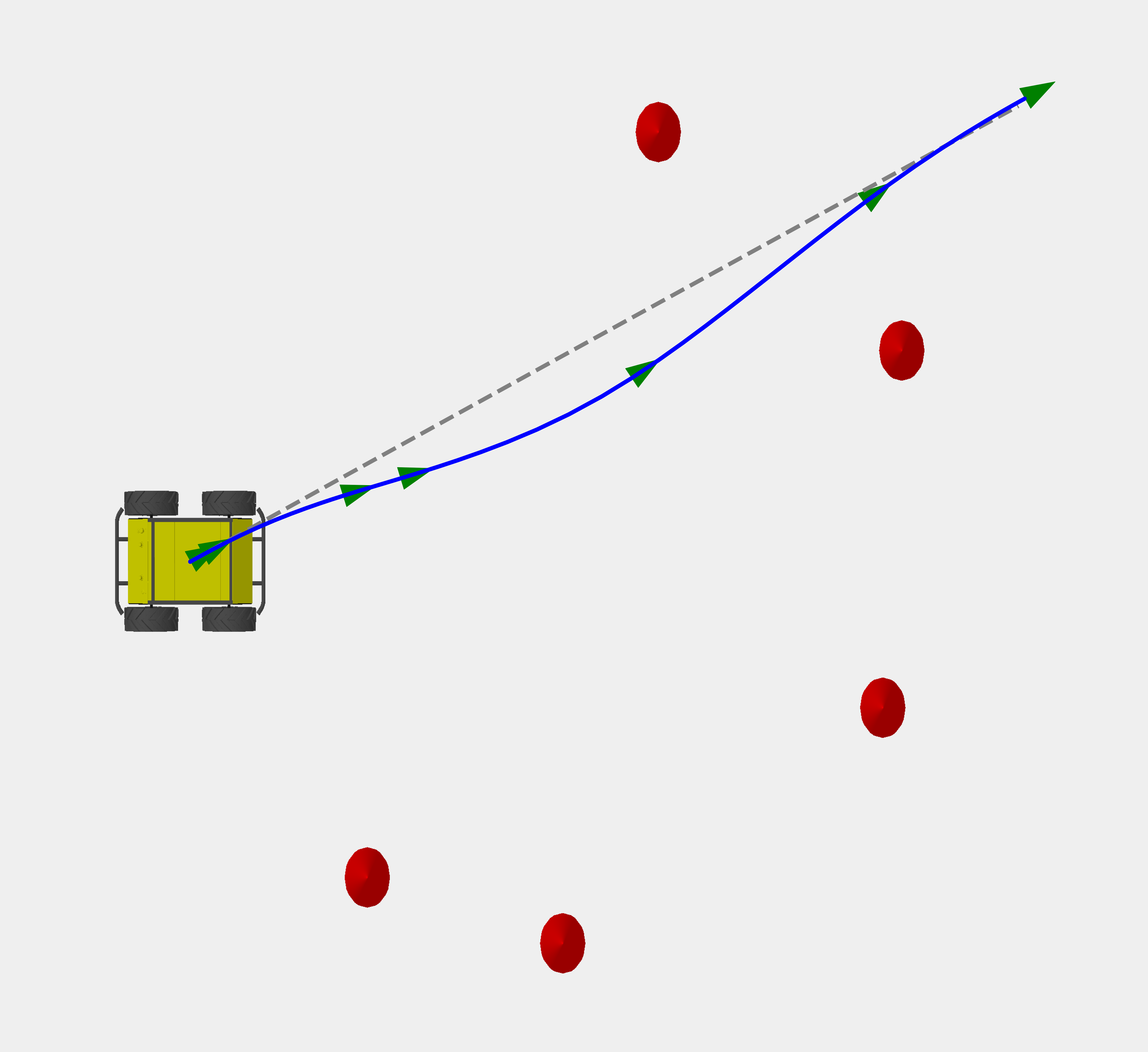}
        \caption{Optimal trajectory generated by TOAST using an LSTM NN predicted constraint-informed warm start (in grey) to solve the MPC problem for an optimal trajectory (in blue). Green arrows indicate the rover's heading, and obstacles are red circles.}
        \label{fig:rover_trajectory}
\end{figure}

\begin{figure}[ht]
        \centering
        \includegraphics[width=.7\linewidth]{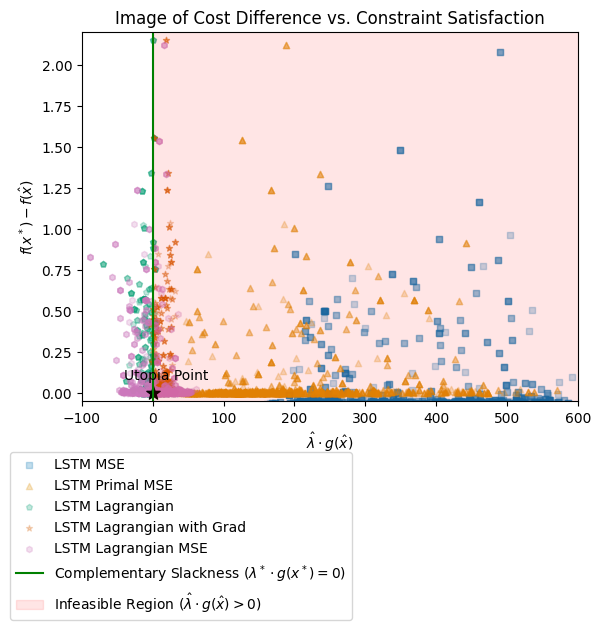}
        \caption{Cost difference vs. constraint satisfaction for~\toast{} predictions on test data.}
        \label{fig:lstm_cost_vs_constraints}
\end{figure}

  \begin{figure}[ht]
        \centering
        \includegraphics[width=.8\linewidth]{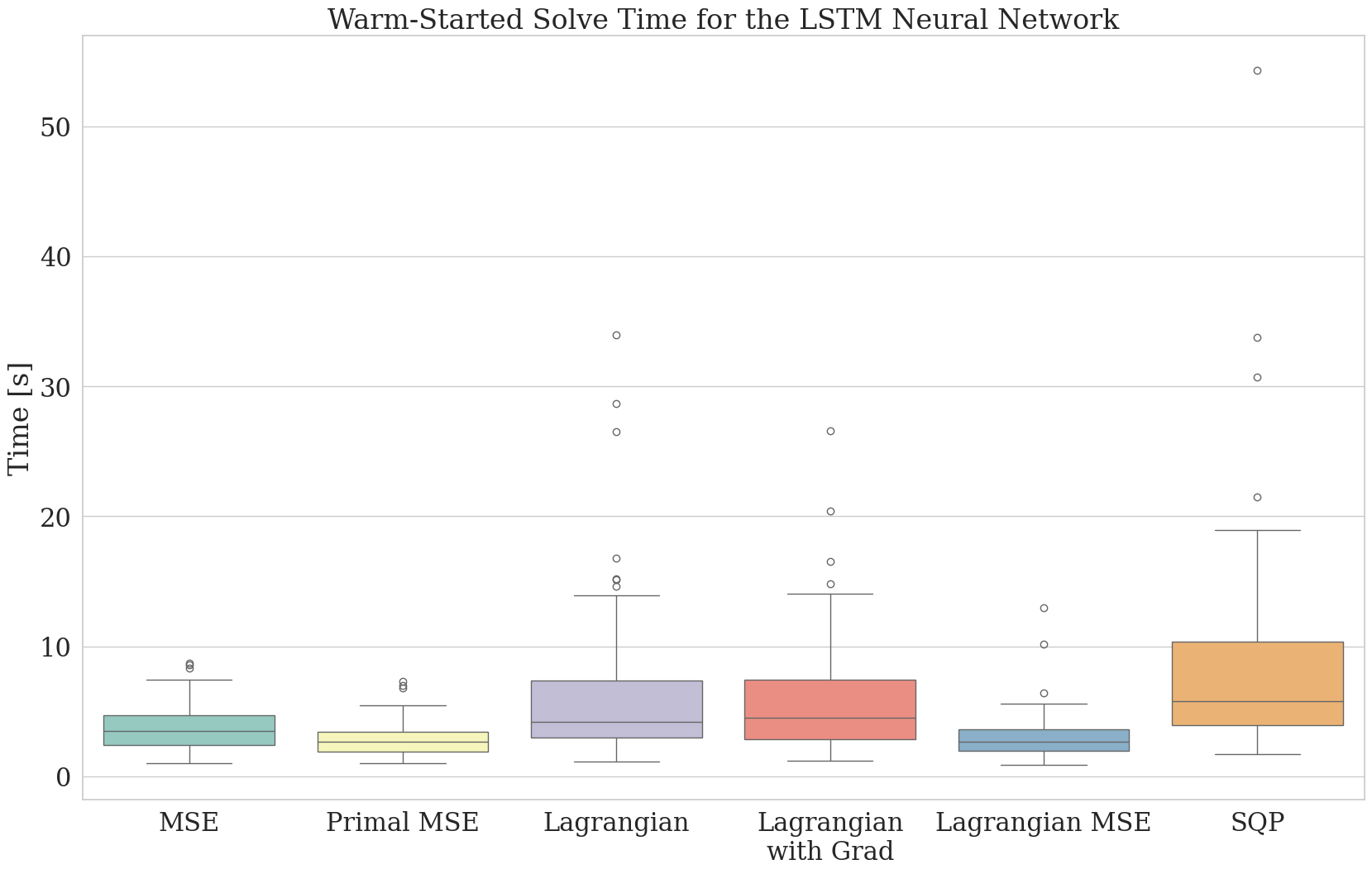}
        \caption{Computation time for test dataset warm-starts using~\toast{}. The constraint-informed LSTM NN provides more than a 5-second speedup from the SQP.}
        \label{fig:timing}
\end{figure}

To apply~\toast{} for the Lunar Rover MPC optimal control problem, an RNN-based LSTM neural network architecture was formulated. We sampled a training dataset of 7200 samples with problem parameters (initial and goal states and five obstacles) sampled from Eqn.~\eqref{eq:lunar rover ocp} with $N=61$. Obstacles were generated along the heading and cross-track, defined by the randomly generated start and goal states. The train-test split for this problem is $80:20$. Additional results, which include a Feedforward NN architecture, are included in the Appendix.

Figure \ref{fig:rover_trajectory} illustrates the lunar rover's trajectory for motion planning around obstacles. TOAST generates an optimal and constraint-satisfying trajectory by first computing an initial guess, shown by the dotted grey line, and then solving the trajectory optimization problem with IPOPT, resulting in the solid blue trajectory.

To analyze the tradeoff between cost minimization and constraint satisfaction, the image of the cost difference vs. constraint satisfaction is plotted in Figures~\ref{fig:lstm_cost_vs_constraints}-\ref{fig:ff_cost_vs_constraints}. From the KKT conditions, the optimal solution meets the complementary slackness condition, shown as the vertical green line. The red region on each plot indicates the location of infeasible dual variable predictions ($\hat{\lambda} < 0$). Lastly, note that predictions of the dual variables computed from the Primal MSE loss are random since they are not included in the loss function computation.

We see immediately that, while close in cost prediction, primal MSE predictions are largely infeasible for the LSTM NN.
Lagrangian loss-trained predictions demonstrate the closest predictions to meeting the complementary slackness condition, with Lagrangian with Grad and Lagrangian MSE close in constraint satisfaction.

Timing results from applying~\toast{} to the online inference problem are plotted in Figure~\ref{fig:timing}. Overall, the Lagrangian-regularized MSE merit function and the Primal MSE are close in dominating performance. Primal MSE is just 190 ms less than the mean computation time for Lagrangian MSE for the LSTM architecture (2.78 vs. 2.97 seconds).

Compared to the full SQP, all merit functions in the~\toast{} software architecture provide warm-starts that significantly reduce computation time. The decision-focused merit functions offer up to a 63\% reduction in mean runtime for the LSTM NN.

Benchmarking against the MSE-based and collision-penalizing spacecraft swarm trajectory planning problem for 10 spacecraft, both warm start techniques reduce the 20-second mean solve time to less than 5 seconds~\citep{SabolYunEtAl2022}. Our implementation using 50 more timesteps and less than 8000 training samples has 2.97 and 4-second runtimes. Which we conclude are comparable with the mean warm-started runtimes provided by~\citep{SabolYunEtAl2022}.

Table~\ref{tab:performance_metrics} shows the performance metrics for each loss function for \toast{}, where CV is the percent of violated constraints, AD is the average degree of constraint violation, and AD is the mean $\pm$, the standard deviation. The Lagrangian with Grad NN achieves the least percent of violated constraints, $11.21\%$, and the smallest degree of violation, $8.59$, for the LSTM NN. Benchmarking against the NN architectures in~\citep{SabolYunEtAl2022}, the average number of collisions increases by 0.043-0.114 for the FF NNs and increases by 0.092 for one of the LSTM NNs when collision-penalization is applied. In contrast,~\toast{} reliably decreases constraint violation by $0.65\%-5.33\%$ for the LSTM NN compared to vanilla~\ac{MSE} loss. Therefore, we have shown that decision-focused merit functions effectively learn trajectory predictions and feasibility.

\begin{table}[ht]
\centering
\caption{Performance Metrics}
\label{tab:performance_metrics}
\begin{tabular}{lll}
\hline
\textbf{Metric} & \textbf{Category} & \textbf{TOAST} \\
\hline
MSE & CV (\%) / AD & 16.54 / 22.95 \(\pm\) 2.12 \\
    & MSE (State / Control) & 3.53 / 0.032 \\
\hline
Primal MSE & CV (\%) / AD & 13.85 / 14.07 \(\pm\) 5.98 \\
           & MSE (State / Control) & 0.068 / 0.033 \\
\hline
Lagrangian & CV (\%) / AD & 15.89 / 11.00 \(\pm\) 5.95 \\
           & MSE (State / Control) & 1.06 / 0.033 \\
\hline
Lagrangian w/ Grad & CV (\%) / AD & \textbf{11.21} / \textbf{8.59} \(\pm\) 2.54 \\
                   & MSE (State / Control) & 0.809 / \textbf{0.032} \\
\hline
Lagrangian MSE & CV (\%) / AD & 13.90 / 13.61 \(\pm\) 5.95 \\
               & MSE (State / Control) & \textbf{0.068} / 0.032 \\
\hline
\end{tabular}
\end{table}

\subsection{Powered Descent Guidance Problem}~\label{subsec:powered_descent}

The powered descent guidance OCP dynamics are modeled in 3 degrees of freedom (DoF), with the spacecraft modeled as a point mass. Equation \ref{eq:zdot_pdg} shows the 3 DoF non-convex dynamics:

\begin{equation} \label{eq:zdot_pdg}
\begin{split}
\dot{z} = 
= \begin{pmatrix}
v \\
\frac{T}{m} - g \\
-\alpha \left\| T \right\|_2.
\end{pmatrix}
\end{split}
\quad\quad
\begin{split}
z = (x, y, z, v_x, v_y, v_z, m)^T,\\
u = (T_x, T_y, T_z)^T.
\end{split}
\end{equation}

The state $z \in  \reals^7$ includes the spacecraft's position, velocity, and mass in a fixed world frame. The control $u \in \reals^3$ includes the thrust control input $T$ in the $x$, $y$, and $z$ directions. The discrete-time update equation is approximated using a backward Euler rule.

The full powered descent guidance OCP minimizes fuel consumption subject to the state dynamics in Equation \ref{eq:zdot_pdg}, control, and state constraints:

\begin{equation} \label{eq:pdg_OCP}
\begin{array}{ll}
\underset{z_{0:N},u_{0:N}}{\textrm{minimize}} \!\!\!& \sum_{t=0}^{N-1} u_t^T u_t \Delta t / \omega_u \\
\text{subject to}\!\!\!& z_0 = z_\textrm{init}, \\
& z_{t+1} \leq z_t + \Delta t \cdot f(z_t, u_t), \quad t = 0, \dots, N-1, \\
& z_{t+1} \geq z_t + \Delta t \cdot f(z_t, u_t), \quad t = 0, \dots, N-1, \\
& z_{\min} \leq z_t \leq z_{\max}, \quad\;\;\;\, t = 0, \dots, N, \\
& u_{\min} \leq u_t \leq u_{\max}, \quad\;\;\;\, t = 0, \dots, N-1, \\
& \| v \|_2 \leq v_{\text{max}}, \quad\;\;\;\, t = 0, \dots, N, \\
& \rho_{\min} \leq \| T \|_2 \leq \rho_{\max}, \quad\;\;\;\, t = 0, \dots, N-1, \\
\end{array}
\end{equation}
where $\omega_u = 10^6$ is a scaling factor and the dynamics equality constraint is decomposed into two inequality constraints. Other constraints include state and control upper and lower bounds, as well as maximum velocity and minimum and maximum thrust constraints.

\subsection{Application: Mars 3 Degree-of-Freedom Powered Descent Guidance}

A transformer-based TOAST neural network architecture to warm start Equation \ref{eq:pdg_OCP}. The training dataset was constructed using 7168 normally distributed samples for the initial conditions $z_0$, as shown in Table \ref{tab:sample_pdg}, with $N = 21$ discretization nodes with the origin as the landing location. The train-test split for this problem is 80:20, resulting in 1792 test samples.

\begin{table}[ht]
\centering
\caption{Sample Ranges for TOAST Training}
    \begin{tabular}{ll}
    \hline
    \textbf{Sampled Variable} & \textbf{Range} \\
    \hline
     $x_0$  & [-2500 m, 2500 m]  \\
     \hline

     $y_0$  & [-2500 m, 2500 m]  \\
     \hline

     $z_0$  & [500 m, 2000 m]  \\
     \hline

     $v_{x0}$  & [-80 m/s, 80 m/s]  \\
     \hline

     $v_{y0}$  & [-80 m/s, 80 m/s]  \\
     \hline

     $v_{z0}$  & [-100 m/s, 0 m/s]  \\
     \hline

     $m_{\text{wet}}$ & [2200 kg, 6600 kg]  \\
     \hline

    \end{tabular}
    
    \label{tab:sample_pdg}
\end{table}

The NN features, consisting of the initial state, and the decision variable targets, the full state and control solutions, are scaled using a min max scaling class to ensure training remains stable. Scaling is achieved by fitting a scale and min value defined by $\text{scale} = \frac{1}{X_{\max} - X_{\min} + 1e-8}$ and $\min = - X_{\min} \times \text{scale}$. Then the scaled value is defined as $X_\text{scaled} = \text{scale} \times X + \min$, which is then clamped to ensure the scaled values remain between zero and one.

\begin{figure}[ht]
    \centering
    \begin{subfigure}[b]{0.4\linewidth}
        \centering
        \includegraphics[width=\linewidth]{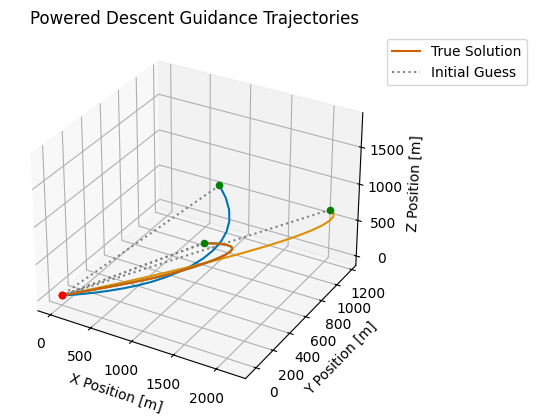}
        \caption{Straight line initial guess and the converged solution for the powered descent guidance optimal control problem.}
        \label{fig:pdg_straightline}
    \end{subfigure}
    \hfill
    \begin{subfigure}[b]{0.5\linewidth}
        \centering
        \includegraphics[width=\linewidth]{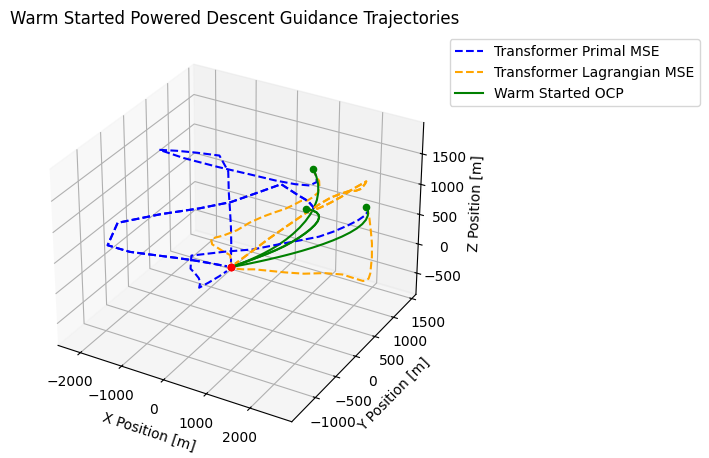}
        \caption{Primal and TOAST Lagrangian MSE trained transformer neural network outputs and the converged warm-started trajectories.}
        \label{fig:pdg_warmstarted}
    \end{subfigure}
    \caption{Comparison of straight line initial guesses and TOAST neural network initial guesses resulting in the converged solutions for the powered descent guidance problem.}
    \label{fig:pdg_comparison}
\end{figure}

Since convergence is not achieved when using IPOPT to solve the OCP in Equation \ref{eq:pdg_OCP} without an initial guess, a straight line initial guess, with thrust predicted to be constant and equal to the average of the minimum and maximum thrust values, was used to obtain both the training and test dataset but also as the baseline comparison for the SQP. Figure \ref{fig:pdg_straightline} shows three straight line initial guesses and the optimal trajectories obtained after warm-starting and solving. Figure \ref{fig:pdg_warmstarted} shows the Primal MSE and TOAST Lagrangian MSE predictions for the same three test cases and the converged trajectories after warm-starting.

Overall, the constraint-informed TOAST Lagrangian MSE trajectories are closer to the locally optimal solutions achieved after solving the OCP. Further, we note that these trajectories are generated by propagating neural network predicted thrusts through the system dynamics in Equation \ref{eq:zdot_pdg}. Therefore, the neural network predicted control inputs are often much closer to the locally optimal control input when compared to constant thrust initial guess used with the straight line initialization.

\begin{figure}[ht]
        \centering
        \includegraphics[width=.6\linewidth]{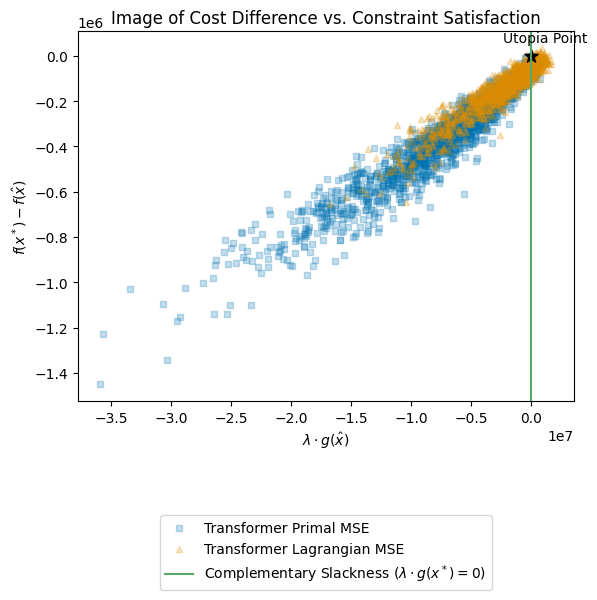}
        \caption{Cost difference vs. constraint satisfaction for \textbf{transformer predictions} on test data.}
        \label{fig:costvconstraints_pdg}
\end{figure}

The tradeoff between cost minimization and constraint satisfaction for the non-standardized powered descent guidance problem is shown in Figure \ref{fig:costvconstraints_pdg}.

The y-axis shows the difference in cost between the optimal cost and the cost computed with the neural network predicted state, control, and dual variables. From the plot, the predicted cost is mostly either equal to or less than the optimal cost. Dual variables for the computation of constraint satisfaction are clipped between the values of -1 and 1, as for the previous problem, so any large values on the x-axis for constraint satisfaction are due to inequality constraint evaluations. Overall, the powered descent guidance problem has a much larger range of values for training and test trajectories, resulting in much larger evaluations of the cost and degree of constraint satisfaction. When comparing the baseline case of Primal MSE to the TOAST architectures Lagrangian MSE, it is evident that there is increased constraint satisfaction as well as closer to optimal cost for Lagrangian MSE.

Table \ref{tab:performance_metricsTransformer} shows the performance metrics for each loss function, where CV is the percent of violated constraints, AD is the average degree of constraint violation, and AD is shown using the mean $\pm$ the standard deviation. Comparing the unscaled constraint violation percentage and the average degree of constraint violation, TOAST Lagrangian MSE achieves a significant decrease in both, decreasing the percentage of constraint violated by 8\% and reducing the degree of violation by almost 70\%. Compared to Primal MSE, Lagrangian MSE also has a lower standard deviation for constraint violation. Lastly, Lagrangian MSE dominates for both standardized state and control accuracy; the MSE for the state variable is reduced by almost 25\%, and control MSE reduces by 50\% when comparing Lagrangian MSE to Primal MSE.

\begin{table}[ht]
\centering
\caption{Performance Metrics for the Transformer Neural Network}
\label{tab:performance_metricsTransformer}
\begin{tabular}{lll}
\hline
\textbf{Architecture} & \textbf{Metric} & \textbf{Transformer} \\
\hline
Primal MSE & CV (\%) / AD\footnotemark[1] & 34.94 / 1.998 e$^{11}$ \(\pm\) 1.438 e$^{11}$ \\
           & MSE (State / Control) & 0.1717 / 0.5505 \\
\hline

\hline
Lagrangian MSE & CV (\%) / AD\footnotemark[1] & \textbf{26.70} / \textbf{6.020 e$^{10}$} \(\pm\) 6.506 e$^{10}$ \\
               & MSE (State / Control) & \textbf{0.1295} / \textbf{0.2748} \\
\hline
\end{tabular}
\begin{flushleft}
\end{flushleft}
\end{table}

\footnotetext[1]{Unscaled}

\begin{figure}[ht]
        \centering
        \includegraphics[width=\linewidth]{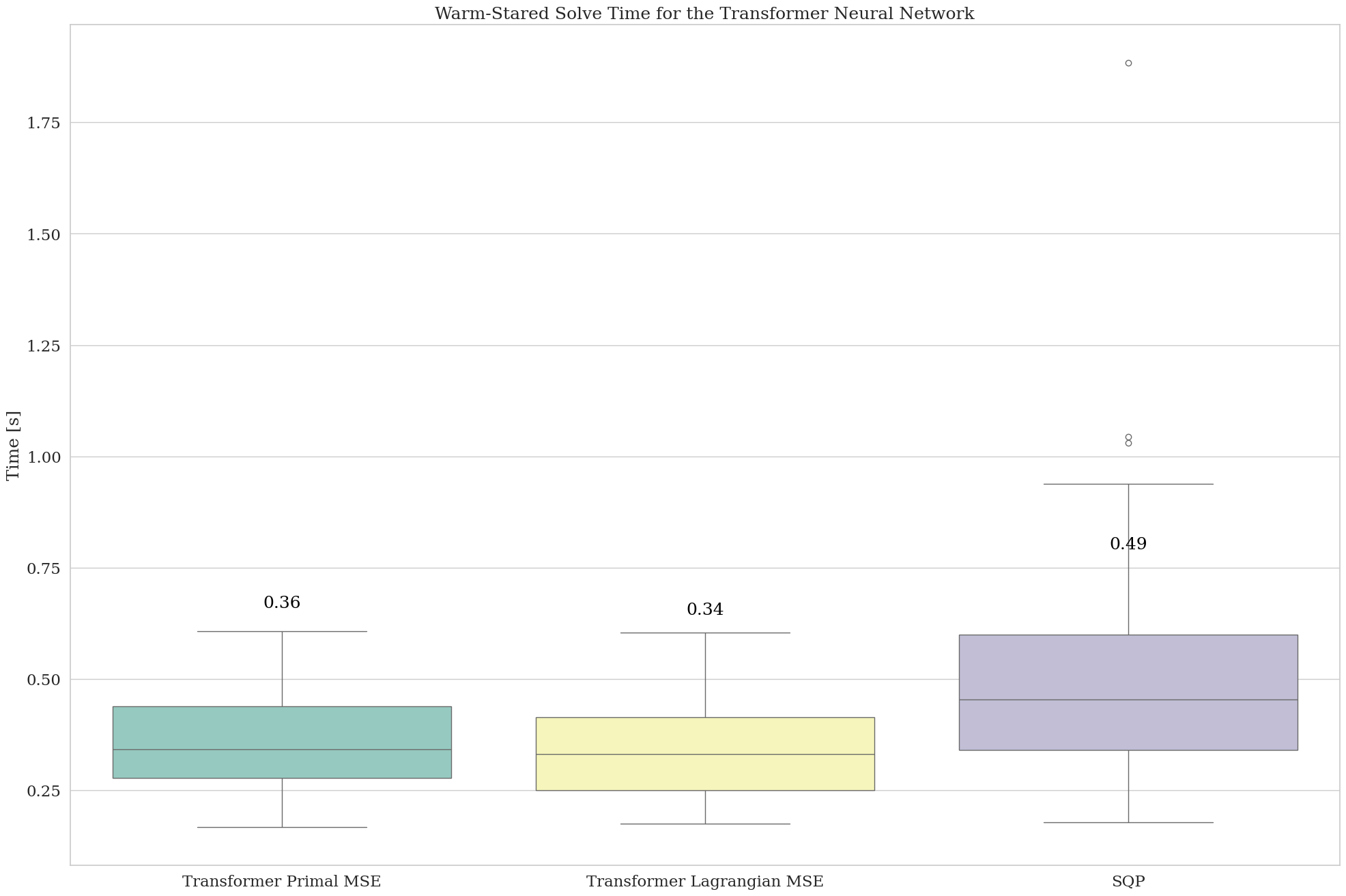}
        \caption{Computation time for 100 test data sampled warm-starts using TOAST Transformer. The constraint-informed Lagrangian MSE merit function provides a 20 ms mean speedup when compared to primal MSE.}
        \label{fig:time_all_pdg}
\end{figure}

Timing results for the TOAST Lagrangian MSE merit function, when compared to Primal MSE, are shown in Figure \ref{fig:time_all_pdg}.

From the timing results, it is evident that compared to the straight-line warm-started SQP, neural network-provided warm starts consistently result in faster runtimes; in the powered descent guidance problem, the SQP runtime was reduced from almost 500 ms to under 360 ms. Constraint-informed warm starting using the TOAST architecture also dominates for computational efficiency, as the Transformer Lagrangian MSE test case reduces runtime by over 30\% when compared to the straight-line warm-started SQP and by more than 5\% when compared to Primal MSE.

\section{Conclusion}\label{sec:conclusion}
By employing a two-step process of offline supervision and online inference using decision-focused merit functions,~\toast{} computes a learned mapping biased towards constraint satisfaction.
Three merit functions were designed for training: Lagrangian Loss, Lagrangian with Gradient Loss, and Lagrangian MSE Loss.
After applying~\toast{} to learn the time-varying policy of Lunar rover MPC and Mars powered descent guidance, benchmarking results demonstrate the expected distributional shifts towards constraint satisfaction on test data and a 100-millisecond to 5-second speedup.
Future work will extend decision-focused learning to the problem of 6-degree-of-freedom powered descent guidance.

\newpage
\appendix
\section*{Appendix}

\begin{figure}[ht]
        \centering
        \includegraphics[width=.7\linewidth]{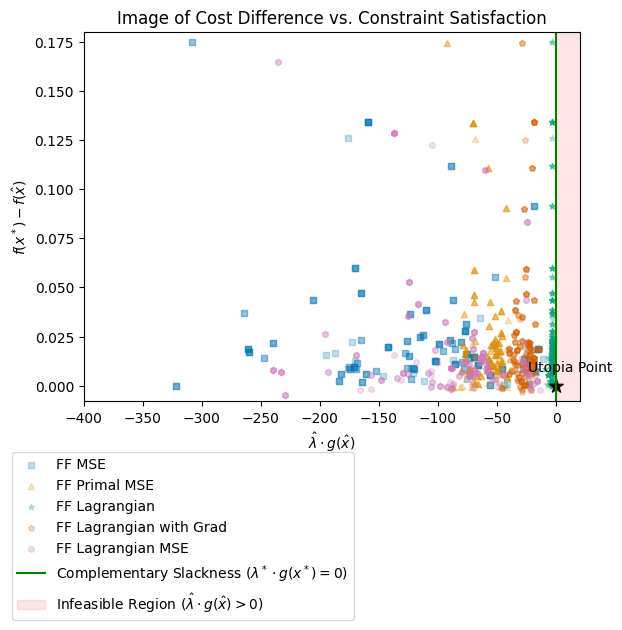}
        \caption{Cost difference vs. constraint satisfaction for \textbf{feedforward predictions} on test data.}
        \label{fig:ff_cost_vs_constraints}
\end{figure}

\begin{figure}[ht]
        \centering
        \includegraphics[width=.8\linewidth]{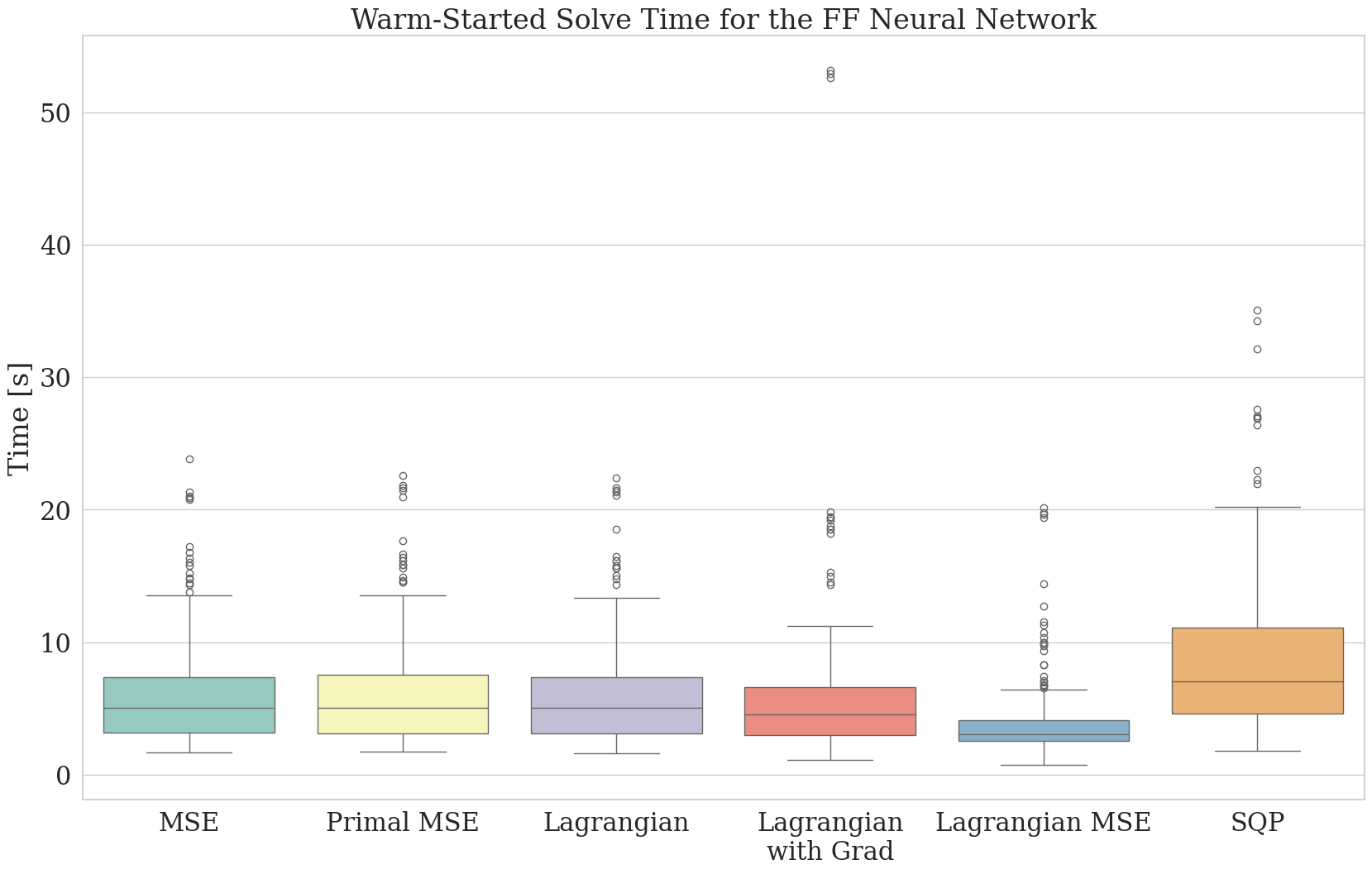}
        \caption{Computation time for test dataset warm-starts using~\toast{} with a feedforward NN. The feedforward NN provides more than a 5-second speedup from the SQP.}
        \label{fig:ff_timing}
\end{figure}

We sampled a training dataset of 1200 samples for the feedforward NN, with problem parameters (initial and goal states and five obstacles) sampled from Eqn.~\eqref{eq:lunar rover ocp} with $N=61$. Obstacles were generated along the heading and cross-track, defined by the randomly generated start and goal states. The train-test split for this problem is $80:20$.

\begin{table}[ht]
\centering
\caption{Performance Metrics for the Feedforward Neural Network}
\label{tab:performance_metricsFF}
\begin{tabular}{lll}
\hline
\textbf{Architecture} & \textbf{Metric} & \textbf{Feedforward} \\
\hline
MSE & CV (\%) / AD & 9.42 / 17.95 \(\pm\) 5.47 \\
    & MSE (State / Control) & 0.157 / 0.00093 \\
\hline
Primal MSE & CV (\%) / AD & 12.56 / 15.73 \(\pm\) 5.04 \\
           & MSE (State / Control) & \textbf{0.122} / 0.005 \\
\hline
Lagrangian & CV (\%) / AD & 3.75 / 11.79 \(\pm\) 3.42 \\
           & MSE (State / Control) & 0.981 / 0.0035 \\
\hline
Lagrangian w/ Grad & CV (\%) / AD & 2.21 / 11.75 \(\pm\) 3.48 \\
                   & MSE (State / Control) & 0.973 / 0.0008 \\
\hline
Lagrangian MSE & CV (\%) / AD & \textbf{1.21} / \textbf{11.66} \(\pm\) 3.37 \\
               & MSE (State / Control) & 1.01 / \textbf{0.00079} \\
\hline
\end{tabular}
\end{table}

Full metrics for the feedforward architecture are shown in Table \ref{tab:performance_metricsFF}. While the feedforward NN denotes the Primal MSE loss as the merit function with the least state error, Lagrangian MSE achieves the minimum state error.

Figure \ref{fig:ff_cost_vs_constraints} shows the cost vs. constraint satisfaction for the feedforward NN architecture. For the feedforward NN, Lagrangian MSE dominates in constraint satisfaction, achieving $1.21\%$ violated constraints and an $11.66$ degree of constraint violation at the cost of a larger state error. Benchmarking against the NN architectures in~\citep{SabolYunEtAl2022}, the average number of collisions increases by 0.043-0.114 for the FF NNs and increases by 0.092 for one of the LSTM NNs when collision-penalization is applied. In contrast,~\toast{} reliably decreases constraint violation by $5.67\%-8.21\%$ for the Feedforward NN, compared to vanilla~\ac{MSE} loss.

Figure \ref{fig:ff_timing} shows the timing results for the feedforward NN. For the feedforward architecture, Lagrangian MSE offers a more than 2-second improvement in mean computation time over MSE (4 vs. 6.22 seconds). Further, the decision-focused merit functions offer up to a 63\% reduction in mean runtime for the LSTM NN and a 54\% reduction in mean runtime for the feedforward NN.

\section*{Acknowledgments}
The authors would like to thank Breanna Johnson and Dan Scharf for their discussions during the development of this work. This research was carried out at the Jet Propulsion Laboratory, California Institute of Technology, under a contract with the National Aeronautics and Space Administration and funded through the internal Research and Technology Development program. This work was supported in part by a NASA Space Technology Graduate Research Opportunity 80NSSC21K1301.

\bibliography{main, project}

\begin{thebibliography}{48}
\newcommand{\enquote}[1]{``#1''}
\providecommand{\natexlab}[1]{#1}
\providecommand{\url}[1]{\texttt{#1}}
\providecommand{\urlprefix}{URL }
\expandafter\ifx\csname urlstyle\endcsname\relax
  \providecommand{\doi}[1]{\discretionary{}{}{}https://doi.org/#1}\else
  \providecommand{\doi}[1]{\discretionary{}{}{}\urlstyle{rm}\url{https://doi.org/#1}}\fi

\bibitem[{Rankin et~al.(2020)Rankin, Maimone, Biesiadecki, Patel, Levine, and Toupet}]{RankinMaimoneEtAl2020}
Rankin, A., Maimone, M., Biesiadecki, J., Patel, N., Levine, D., and Toupet, O., \enquote{Driving {Curiosity}: {Mars} {Rover} Mobility Trends During the First Seven Years,} \emph{{IEEE Aerospace Conference}}, 2020.

\bibitem[{Verma et~al.(2023)Verma, Maimone, Gaines, Francis, Estlin, Kuhn, Rabideau, Chien, {McHenry}, Graser, Rankin, and Thiel}]{VermaMaimoneEtAl2023}
Verma, V., Maimone, M.~W., Gaines, D.~M., Francis, R., Estlin, T.~A., Kuhn, S.~R., Rabideau, G.~R., Chien, S.~A., {McHenry}, M., Graser, E.~J., Rankin, A.~L., and Thiel, E.~R., \enquote{Autonomous robotics is driving {Perseverance} rover’s progress on {Mars},} \emph{{Science Robotics}}, Vol.~8, No.~80, 2023, pp. 1--12.

\bibitem[{{National Academies of Sciences, Engineering, and Medicine}(2022)}]{NASEM2022}
{National Academies of Sciences, Engineering, and Medicine}, \enquote{Origins, Worlds, and Life: A Decadal Strategy for Planetary Science and Astrobiology 2023--2032.} Tech. rep., {National Academy Press}, 2022.

\bibitem[{Keane et~al.(2022)Keane, Tikoo, and Elliott}]{KeaneTikooEtAl2022}
Keane, J.~T., Tikoo, S.~M., and Elliott, J., \enquote{{Endurance}: {Lunar} {South} {Pole}-{Atken} {Basin} Traverse and Sample Return Rover,} Tech. rep., {National Academy Press}, 2022.

\bibitem[{Toupet et~al.(2020)Toupet, {Del}~{Sesto}, Ono, Myint, {Vander}~{Hook}, and {McHenry}}]{ToupetDelSestoEtAl2020}
Toupet, O., {Del}~{Sesto}, T., Ono, M., Myint, S., {Vander}~{Hook}, J., and {McHenry}, M., \enquote{A {ROS}-based Simulator for Testing the {Enhanced} {Autonomous} {Navigation} of the {Mars} 2020 {Rover},} \emph{{IEEE Aerospace Conference}}, 2020.

\bibitem[{Daftry et~al.(2022)Daftry, Abcouwer, {Del}~{Sesto}, Venkatraman, Song, Igel, Byon, Rosolia, Yue, and Ono}]{DaftryAbcouwerEtAl2022}
Daftry, S., Abcouwer, N., {Del}~{Sesto}, T., Venkatraman, S., Song, J., Igel, L., Byon, A., Rosolia, U., Yue, Y., and Ono, M., \enquote{{MLNav}: Learning to Safely Navigate on {Martian} Terrains,} \emph{{IEEE Robotics and Automation Letters}}, Vol.~7, No.~2, 2022, pp. 5461--5468.

\bibitem[{Dueri et~al.(2017)Dueri, Acikmese, Scharf, and Harris}]{Dueri2017}
Dueri, D., Acikmese, B., Scharf, D.~P., and Harris, M.~W., \enquote{Customized Real-Time Interior-Point Methods for Onboard Powered-Descent Guidance,} \emph{JGCD Special Issue on Computational Guidance and Control}, Vol.~40, No.~2, 2017.
\newblock \doi{10.2514/1.G001480}.

\bibitem[{Elango et~al.(2022)Elango, Kamath, Yu, Acikmese, Mesbahi, and Carson}]{Elango2022}
Elango, P., Kamath, A.~G., Yu, Y., Acikmese, B., Mesbahi, M., and Carson, J.~M., \enquote{A Customized First-Order Solver for Real-Time Powered-Descent Guidance,} \emph{AIAA SCITECH}, 2022.
\newblock \doi{10.2514/6.2022-0951}.

\bibitem[{Kamath et~al.(2023)Kamath, Elango, Kim, Mceowen, Yu, Carson, Mesbahi, and Acikmese}]{kamath2023customized}
Kamath, A.~G., Elango, P., Kim, T., Mceowen, S., Yu, Y., Carson, J.~M., Mesbahi, M., and Acikmese, B., \enquote{Customized Real-Time First-Order Methods for Onboard Dual Quaternion-based 6-DoF Powered-Descent Guidance,} \emph{AIAA 2023-2003, Session: Entry, Descent and Landing GN\&C Technology V}, AIAA, 2023.
\newblock \doi{10.2514/6.2023-2003}, \urlprefix\url{https://doi.org/10.2514/6.2023-2003}.

\bibitem[{Kamath et~al.(2022)Kamath, Elango, Yu, Mceowen, Chari, Carson~III, and Açıkmeşe}]{kamath2022realtime}
Kamath, A.~G., Elango, P., Yu, Y., Mceowen, S., Chari, G.~M., Carson~III, J.~M., and Açıkmeşe, B., \enquote{Real-Time Sequential Conic Optimization for Multi-Phase Rocket Landing Guidance,} \emph{arXiv preprint arXiv:2212.00375}, 2022.
\newblock \doi{10.48550/arXiv.2212.00375}, \urlprefix\url{https://doi.org/10.48550/arXiv.2212.00375}.

\bibitem[{Eren et~al.(2017)Eren, Prach, Ko{\c{c}}er, Rakovi{\'c}, Kayacan, and A{\c{c}}ikmese}]{ErenPrachEtAl2017}
Eren, U., Prach, A., Ko{\c{c}}er, B.~B., Rakovi{\'c}, S.~V., Kayacan, E., and A{\c{c}}ikmese, B., \enquote{Model Predictive Control in Aerospace Systems: Current State and Opportunities,} \emph{{AIAA Journal of Guidance, Control, and Dynamics}}, Vol.~40, No.~7, 2017, pp. 1541--1566.

\bibitem[{Amos(2023)}]{Amos2023}
Amos, B., \enquote{Tutorial on Amortized Optimization,} \emph{{Foundations and Trends in Machine Learning}}, Vol.~16, No.~5, 2023, p. 732.

\bibitem[{Kim et~al.(2018)Kim, Wiseman, Miller, Sontag, and Rush}]{KimWisemanEtAl2018}
Kim, Y., Wiseman, S., Miller, A., Sontag, D., and Rush, A., \enquote{Semi-Amortized Variational Autoencoders,} \emph{{Int.\ Conf.\ on Machine Learning}}, 2018.

\bibitem[{Marino et~al.(2018)Marino, Yue, and Mandt}]{MarinoYueEtAl2018}
Marino, J., Yue, Y., and Mandt, S., \enquote{Iterative Amortized Inference,} \emph{{Int.\ Conf.\ on Machine Learning}}, 2018.

\bibitem[{Donti et~al.(2021)Donti, Rolnick, and Kolter}]{DontiRolnickEtAl2021}
Donti, P., Rolnick, D., and Kolter, J.~Z., \enquote{{DC3}: A learning method for optimization with hard constraints,} \emph{{Int.\ Conf.\ on Learning Representations}}, 2021.

\bibitem[{Kotary et~al.(2021)Kotary, Fioretto, {Van}~{Hentenryck}, and Wilder}]{KotaryFiorettoEtAl2021}
Kotary, J., Fioretto, F., {Van}~{Hentenryck}, P., and Wilder, B., \enquote{End-to-End Constrained Optimization Learning: A Survey,} \emph{{Int.\ Joint Conf.\ on Artificial Intelligence}}, 2021.

\bibitem[{Cauligi et~al.(2022{\natexlab{a}})Cauligi, Culbertson, Schmerling, Schwager, Stellato, and Pavone}]{CauligiCulbertsonEtAl2022}
Cauligi, A., Culbertson, P., Schmerling, E., Schwager, M., Stellato, B., and Pavone, M., \enquote{{CoCo}: Online Mixed-Integer Control via Supervised Learning,} \emph{{IEEE Robotics and Automation Letters}}, Vol.~7, No.~2, 2022{\natexlab{a}}, pp. 1447--1454.

\bibitem[{Dua et~al.(2008)Dua, Kouramas, Dua, and Pistikopoulos}]{DuaKouramasEtAl2008}
Dua, P., Kouramas, K., Dua, V., and Pistikopoulos, E.~N., \enquote{{MPC} on a chip - Recent advances on the application of multi-parametric model-based control,} \emph{{Computers \& Chemical Engineering}}, Vol.~32, No. 4-5, 2008, pp. 754--765.

\bibitem[{{de} {la}~{Croix} et~al.(2024){de} {la}~{Croix}, Rossi, Brockers, Aguilar, Albee, Boroson, Cauligi, Delaune, and {others}}]{DeLaCroixRossiEtAl2024}
{de} {la}~{Croix}, J.-P., Rossi, F., Brockers, R., Aguilar, D., Albee, K., Boroson, E., Cauligi, A., Delaune, J., and {others}, \enquote{Multi-Agent Autonomy for Space Exploration on the {CADRE} {Lunar} Technology Demonstration Mission,} \emph{{IEEE Aerospace Conference}}, 2024.

\bibitem[{Ghosh et~al.(2024)Ghosh, Tomar, Mhatre, Sumithra, Kumar, and Siva}]{GhoshTomarEtAl2024}
Ghosh, R., Tomar, S., Mhatre, C.~S., Sumithra, K., Kumar, G. V. P.~B., and Siva, M.~S., \enquote{Path Planning for the {Pragyan} Rover: Experien and Challenges,} \emph{{IEEE Int.\ Conf.\ on Space Robotics}}, 2024.

\bibitem[{Finn et~al.(2017)Finn, Abbeel, and Levine}]{FinnAbbeelEtAl2017}
Finn, C., Abbeel, P., and Levine, S., \enquote{Model-agnostic meta-learning for fast adaptation of deep networks,} \emph{{Int.\ Conf.\ on Machine Learning}}, 2017.

\bibitem[{Andrychowicz et~al.(2016)Andrychowicz, Denil, Colmenarejo, Hoffman, Pfau, Schaul, Shillingford, and {de}~{Freitas}}]{AndrychowiczDenilEtAl2016}
Andrychowicz, M., Denil, M., Colmenarejo, S.~G., Hoffman, M.~W., Pfau, D., Schaul, T., Shillingford, B., and {de}~{Freitas}, N., \enquote{Learning to learn by gradient descent by gradient descent,} \emph{{Conf.\ on Neural Information Processing Systems}}, 2016.

\bibitem[{Chen et~al.(2022)Chen, Wang, Atanasov, Kumar, and Morari}]{ChenWangEtAl2022}
Chen, S.~W., Wang, T., Atanasov, N., Kumar, V., and Morari, M., \enquote{Large scale model predictive control with neural networks and primal active sets,} \emph{{Automatica}}, Vol. 135, 2022, p. 109947.

\bibitem[{Sambharya et~al.(2023)Sambharya, Hall, Amos, and Stellato}]{SambharyaHallEtAll2022}
Sambharya, R., Hall, G., Amos, B., and Stellato, B., \enquote{End-to-End Learning to Warm-Start for Real-Time Quadratic Optimization,} \emph{{Learning for Dynamics \& Control}}, 2023.

\bibitem[{Morelli et~al.(2024)Morelli, Hofmann, and Topputo}]{MorelliHofmannEtAl2024}
Morelli, A.~C., Hofmann, C., and Topputo, F., \enquote{Warm-Start of Interior-Point Methods Applied to Sequential Convex Programming,} \emph{{IEEE Transactions on Aerospace and Electronic Systems}}, 2024, pp. 1--10.

\bibitem[{Ichnowski et~al.(2020)Ichnowski, Avigal, Satish, and Goldberg}]{IchnowskiAvigalEtAl2020}
Ichnowski, J., Avigal, Y., Satish, V., and Goldberg, K., \enquote{Deep learning can accelerate grasp-optimized motion planning,} \emph{{Science Robotics}}, Vol.~5, No.~48, 2020, pp. 1--12.

\bibitem[{Briden et~al.(2024)Briden, Gurga, Johnson, Cauligi, and Linares}]{BridenGurgaEtAl2024}
Briden, J., Gurga, T., Johnson, B., Cauligi, A., and Linares, R., \enquote{Improving Computational Efficiency for Powered Descent Guidance via Transformer-based Tight Constraint Prediction,} \emph{{AIAA Scitech Forum}}, 2024.

\bibitem[{Guffanti et~al.(2024)Guffanti, Gammelli, {D'Amico}, and Pavone}]{GuffantiGammelliEtAl2024}
Guffanti, T., Gammelli, D., {D'Amico}, S., and Pavone, M., \enquote{Transformers for Trajectory Optimization with Application to Spacecraft Rendezvous,} \emph{{IEEE Aerospace Conference}}, 2024.

\bibitem[{Sambharya and Stellato(2024)}]{SambharyaStellato2024}
Sambharya, R., and Stellato, B., \enquote{Data-Driven Performance Guarantees for Classical and Learned Optimizers,} , 2024.
\newblock {Available at }\url{https://arxiv.org/pdf/2404.13831}.

\bibitem[{Sabol et~al.(2022)Sabol, Yun, Adil, Choi, and Madani}]{SabolYunEtAl2022}
Sabol, A., Yun, K., Adil, M., Choi, C., and Madani, R., \enquote{Machine Learning Based Relative Orbit Transfer for Swarm Spacecraft Motion Planning,} \emph{{IEEE Aerospace Conference}}, 2022.

\bibitem[{Cauligi et~al.(2022{\natexlab{b}})Cauligi, Chakrabarty, {Di}~{Cairano}, and Quirynen}]{CauligiChakrabartyEtAl2022}
Cauligi, A., Chakrabarty, A., {Di}~{Cairano}, S., and Quirynen, R., \enquote{{PRISM}: Recurrent Neural Networks and Presolve Methods for Fast Mixed-integer Optimal Control,} \emph{{Learning for Dynamics \& Control}}, 2022{\natexlab{b}}.

\bibitem[{Vaswani et~al.(2017)Vaswani, Shazeer, Parmar, Uszkoreit, Jones, Gomez, Kaiser, and Polosukhin}]{Vaswani2017}
Vaswani, A., Shazeer, N., Parmar, N., Uszkoreit, J., Jones, L., Gomez, A., Kaiser, L., and Polosukhin, I., \enquote{Attention Is All You Need,} \emph{Conference on Neural Information Processing Systems}, Vol.~30, 2017.

\bibitem[{Wilder et~al.(2019)Wilder, Dilkina, and Tambe}]{WilderDilkinaEtAl2019}
Wilder, B., Dilkina, B., and Tambe, M., \enquote{Melding the Data-Decisions Pipeline: Decision-Focused Learning for Combinatorial Optimization,} \emph{{Proc.\ AAAI Conf.\ on Artificial Intelligence}}, 2019.

\bibitem[{Mandi et~al.(2023)Mandi, Kotary, Berden, Mulamba, Bucarey, Guns, and Fioretto}]{MandiKotaryEtAl2023}
Mandi, J., Kotary, J., Berden, S., Mulamba, M., Bucarey, V., Guns, T., and Fioretto, F., \enquote{Decision-Focused Learning: Foundations, State of the Art, Benchmark and Future Opportunities,} , 2023.
\newblock {Available at }\url{https://arxiv.org/abs/2307.13565}.

\bibitem[{Nocedal and Wright(2006)}]{NocedalWright2006}
Nocedal, J., and Wright, S.~J., \emph{Numerical Optimization}, 2\textsuperscript{nd} ed., {Springer}, 2006.

\bibitem[{Zhang et~al.(2019)Zhang, Bujarbaruah, and Borrelli}]{ZhangBujarbaruahEtAl2019}
Zhang, X., Bujarbaruah, M., and Borrelli, F., \enquote{Safe and Near-Optimal Policy Learning for Model Predictive Control using Primal-Dual Neural Networks,} \emph{{American Control Conference}}, 2019.

\bibitem[{Reske et~al.(2021)Reske, Carius, Ma, Farshidian, and Hutter}]{ReskeCariusEtAl2021}
Reske, A., Carius, J., Ma, Y., Farshidian, F., and Hutter, M., \enquote{Imitation Learning from {MPC} for Quadrupedal Multi-Gait Control,} \emph{{Proc.\ IEEE Conf.\ on Robotics and Automation}}, 2021.

\bibitem[{Tagliabue et~al.(2022)Tagliabue, Kim, Everett, and How}]{TagliabueKimEtAl2022}
Tagliabue, A., Kim, D.-K., Everett, M., and How, J.~P., \enquote{Demonstration-Efficient Guided Policy Search via Imitation of Robust Tube {MPC},} \emph{{Proc.\ IEEE Conf.\ on Robotics and Automation}}, 2022.

\bibitem[{Betts(1998)}]{Betts1998}
Betts, J.~T., \enquote{Survey of Numerical Methods for Trajectory Optimization,} \emph{{AIAA Journal of Guidance, Control, and Dynamics}}, Vol.~21, No.~2, 1998, pp. 193--207.

\bibitem[{Kelly(2017)}]{Kelly2017}
Kelly, M., \enquote{An Introduction to Trajectory Optimization: How to Do Your Own Direct Collocation,} \emph{{SIAM Review}}, Vol.~59, No.~4, 2017, pp. 849 -- 904.

\bibitem[{Boyd and Vandenberghe(2004)}]{BoydVandenberghe2004}
Boyd, S., and Vandenberghe, L., \emph{Convex Optimization}, {Cambridge Univ.\ Press}, 2004.

\bibitem[{Paszke et~al.(2017)Paszke, Gross, Chintala, Chanan, Yang, DeVito, Lin, Desmaison, Antiga, and Lerer}]{PaszkeGrossEtAl2017}
Paszke, A., Gross, S., Chintala, S., Chanan, G., Yang, E., DeVito, Z., Lin, Z., Desmaison, A., Antiga, L., and Lerer, A., \enquote{Automatic differentiation in {PyTorch},} \emph{{Conf.\ on Neural Information Processing Systems - Autodiff Workshop}}, 2017.

\bibitem[{Kingma and Ba(2015)}]{KingmaBa2015}
Kingma, D.~P., and Ba, J.~L., \enquote{Adam: A method for stochastic optimization,} \emph{{Int.\ Conf.\ on Learning Representations}}, 2015.

\bibitem[{Andersson et~al.(2019)Andersson, Gillis, Horn, Rawlings, and Diehl}]{AnderssonGillisEtAl2019}
Andersson, J. A.~E., Gillis, J., Horn, G., Rawlings, J.~B., and Diehl, M., \enquote{{CasADi}: A software framework for nonlinear optimization and optimal control,} \emph{{Mathematical Programming Computation}}, Vol.~11, No.~1, 2019, pp. 1--36.

\bibitem[{W{\"a}chter and Biegler(2006)}]{WachterBiegler2006}
W{\"a}chter, A., and Biegler, L.~T., \enquote{On the implementation of an interior-point filter line-search algorithm for large-scale nonlinear programming,} \emph{{Mathematical Programming}}, Vol. 106, No.~1, 2006, pp. 25--57.

\bibitem[{Haeser et~al.(2021)Haeser, Hinder, and Ye}]{HaeserHinderEtAl2021}
Haeser, G., Hinder, O., and Ye, Y., \enquote{On the behavior of {Lagrange} multipliers in convex and nonconvex infeasible interior point methods,} \emph{{Mathematical Programming}}, Vol. 186, 2021, pp. 257--288.

\bibitem[{Liu et~al.(2018)Liu, Paden, and Ozguner}]{LiuPadenEtAl2018}
Liu, P., Paden, B., and Ozguner, U., \enquote{Model Predictive Trajectory Optimization and Tracking for On-Road Autonomous Vehicles,} \emph{{Proc.\ IEEE Int.\ Conf.\ on Intelligent Transportation Systems}}, 2018.

\bibitem[{Szegedy et~al.(2014)Szegedy, Zaremba, Sutskever, Bruna, Erhan, Goodfellow, and Fergus}]{SzegedyZarembaEtAl2014}
Szegedy, C., Zaremba, W., Sutskever, I., Bruna, J., Erhan, D., Goodfellow, I., and Fergus, R., \enquote{Intriguing properties of neural networks,} , 2014.
\newblock {Available at }\url{https://arxiv.org/abs/1312.6199}.

\end{thebibliography}

\end{document}